\documentclass[letterpaper]{article} % DO NOT CHANGE THIS
\usepackage[draft]{aaai25}  % DO NOT CHANGE THIS
\usepackage{times}  % DO NOT CHANGE THIS
\usepackage{helvet}  % DO NOT CHANGE THIS
\usepackage{courier}  % DO NOT CHANGE THIS
\usepackage[hyphens]{url}  % DO NOT CHANGE THIS
\usepackage{graphicx} % DO NOT CHANGE THIS
\usepackage{natbib}  % DO NOT CHANGE THIS AND DO NOT ADD ANY OPTIONS TO IT
\usepackage{caption} % DO NOT CHANGE THIS AND DO NOT ADD ANY OPTIONS TO IT
\usepackage{algorithm}
\usepackage{algorithmic}
\usepackage{newfloat}
\usepackage{listings}

\usepackage{amsmath} 
\usepackage{bm} 
\usepackage{soul}
\usepackage{multirow}
\usepackage[inline]{enumitem}
\usepackage{comment}
\usepackage{tipa}
\usepackage{longtable}
\usepackage{makecell}
\usepackage{siunitx}
\usepackage{subcaption}
\usepackage{booktabs}
\usepackage{rotating}
\usepackage{titletoc}
\usepackage{placeins} % for \FloatBarrier
\usepackage{amsfonts}
\usepackage[title]{appendix}

\DeclareMathOperator*{\argmin}{arg\,min}


\usepackage{xr}
\makeatletter
\newcommand*{\addFileDependency}[1]{% argument=file name and extension
  \typeout{(#1)}% latexmk will find this if $recorder=0 (however, in that case, it will ignore #1 if it is a .aux or .pdf file etc and it exists! if it doesn't exist, it will appear in the list of dependents regardless)
  \@addtofilelist{#1}% if you want it to appear in \listfiles, not really necessary and latexmk doesn't use this
  \IfFileExists{#1}{}{\typeout{No file #1.}}% latexmk will find this message if #1 doesn't exist (yet)
}

\usepackage{tcolorbox}

\usepackage{xr}
\makeatletter
\DeclareCaptionStyle{ruled}{labelfont=normalfont,labelsep=colon,strut=off} % DO NOT CHANGE THIS
\floatstyle{ruled}
\newfloat{listing}{tb}{lst}{}
\floatname{listing}{Listing}
\title{FedPT: Federated Proxy-Tuning of Large Language Models on Resource-Constrained Edge Devices}

\author{
    Zhidong Gao\textsuperscript{\rm 1}, 
    Yu Zhang\textsuperscript{\rm 1}, 
    Zhenxiao Zhang\textsuperscript{\rm 1}, 
    Yanmin Gong\textsuperscript{\rm 1}, 
    Yuanxiong Guo\textsuperscript{\rm 2}\textsuperscript{\dag}
}
\affiliations{
    \textsuperscript{\rm 1}Department of Electrical and Computer Engineering\\
    \textsuperscript{\rm 2}Department of Information Systems and Cyber Security\\
    The University of Texas at San Antonio, San Antonio, USA\\
    \{zhidong.gao@my., yu.zhang@my., zhenxiao.zhang@my., , yanmin.gong@, yuanxiong.guo@\}utsa.edu
}

\usepackage{bibentry}
\begin{document}

\maketitle
\renewcommand{\thefootnote}{\dag}
\footnotetext{Corresponding Authors}
\begin{abstract}
%Pre-trained large language models (LMs) have demonstrated superior performance across a variety of linguistic tasks. While these models benefit from extensive training data, they often require fine-tuning on specific datasets to effectively address different downstream tasks. 

Despite demonstrating superior performance across a variety of linguistic tasks, pre-trained large language models (LMs) often require fine-tuning on specific datasets to effectively address different downstream tasks. However, fine-tuning these LMs for downstream tasks necessitates collecting data from individuals, which raises significant privacy concerns. Federated learning (FL) has emerged as the de facto solution, enabling collaborative model training without sharing raw data. While promising, federated fine-tuning of large LMs faces significant challenges, including restricted access to model parameters and high computation, communication, and memory overhead. To address these challenges, this paper introduces \textbf{Fed}erated \textbf{P}roxy-\textbf{T}uning (FedPT), a novel framework for federated fine-tuning of black-box large LMs, requiring access only to their predictions over the output vocabulary instead of their parameters. Specifically, devices in FedPT first collaboratively tune a smaller LM, and then the server combines the knowledge learned by the tuned small LM with the knowledge learned by the larger pre-trained LM to construct a large proxy-tuned LM that can reach the performance of directly tuned large LMs. The experimental results demonstrate that FedPT can significantly reduce computation, communication, and memory overhead while maintaining competitive performance compared to directly federated fine-tuning of large LMs. FedPT offers a promising solution for efficient, privacy-preserving fine-tuning of large LMs on resource-constrained devices, broadening the accessibility and applicability of state-of-the-art large LMs.

\end{abstract}

%%%%%%%%%%%%%%%%%%%%%%%%%%%%%%%
\section{Introduction}
%%%%%%%%%%%%%%%%%%%%%%%%%%%%%%%

The emerging large language models (LMs) have demonstrated remarkable zero-shot and few-shot learning capabilities across various language tasks, such as text generation, question-answering, and machine translation. Large LMs, such as LLaMA \cite{touvron2023llama} and GPT-4 \cite{achiam2023gpt}, are trained on massive, diverse, and public datasets with up to hundreds of billion parameters. To adapt a general large LM for a specific task, it is usually fine-tuned on task-oriented datasets to meet the desired quality of service.  For instance, PMC-LLaMA \cite{wu2023pmc} is fine-tuned on medical data to achieve improved accuracy on medical-related questions. In practice, these datasets (e.g., user reviews and emails) are often distributed across devices, and collecting these datasets can be costly and may compromise user privacy.

To overcome this issue, federated learning (FL) \cite{mcmahan2017communication}, which enables collaborative model training without sharing the raw data, is a de facto approach. %The FL training process consists of two iterative phases: local training and global aggregation. %
However, 
%as shown in {\color{red}Fig/table.~X(a)}, 
there are several significant challenges for directly fine-tuning large LMs in FL: 
\begin{enumerate*}[label={\arabic*)}]
    \item \emph{Memory Overhead}. Training large models requires significant memory, often exceeding 10 GB~\cite{wang2023overview,rajbhandari2020zero}. Most devices have RAM capacities of 4-8 GB~\cite{blogdoiphone2024}, which is insufficient for such tasks.
    \item \emph{Computation Overhead}. Even on a GPU-equipped device, local computations can take several hundred seconds per round. Consequently, a fine-tuning session may extend over several days.
    \item \emph{Communication Overhead}. In each FL round, participating devices are required to download the latest global model and then upload their local models. 
%    \item \emph{Black-box Model}. The pre-trained LMs could be proprietary (e.g., GPT-4), and thus their model weights are private, making it impossible to directly tune these models on edge devices in FL.  
%
    %Moreover, it provides an avenue for users to customize proprietary LMs when the output logits are provided, even when weights are not, allowing organizations to keep their pretrained models private while satisfying user needs for adaptation.
    % \hl{which involves transferring hundreds of MBs of data.}
\end{enumerate*}

\begin{figure*}[t]
\centering
  \includegraphics[width=0.9\linewidth]{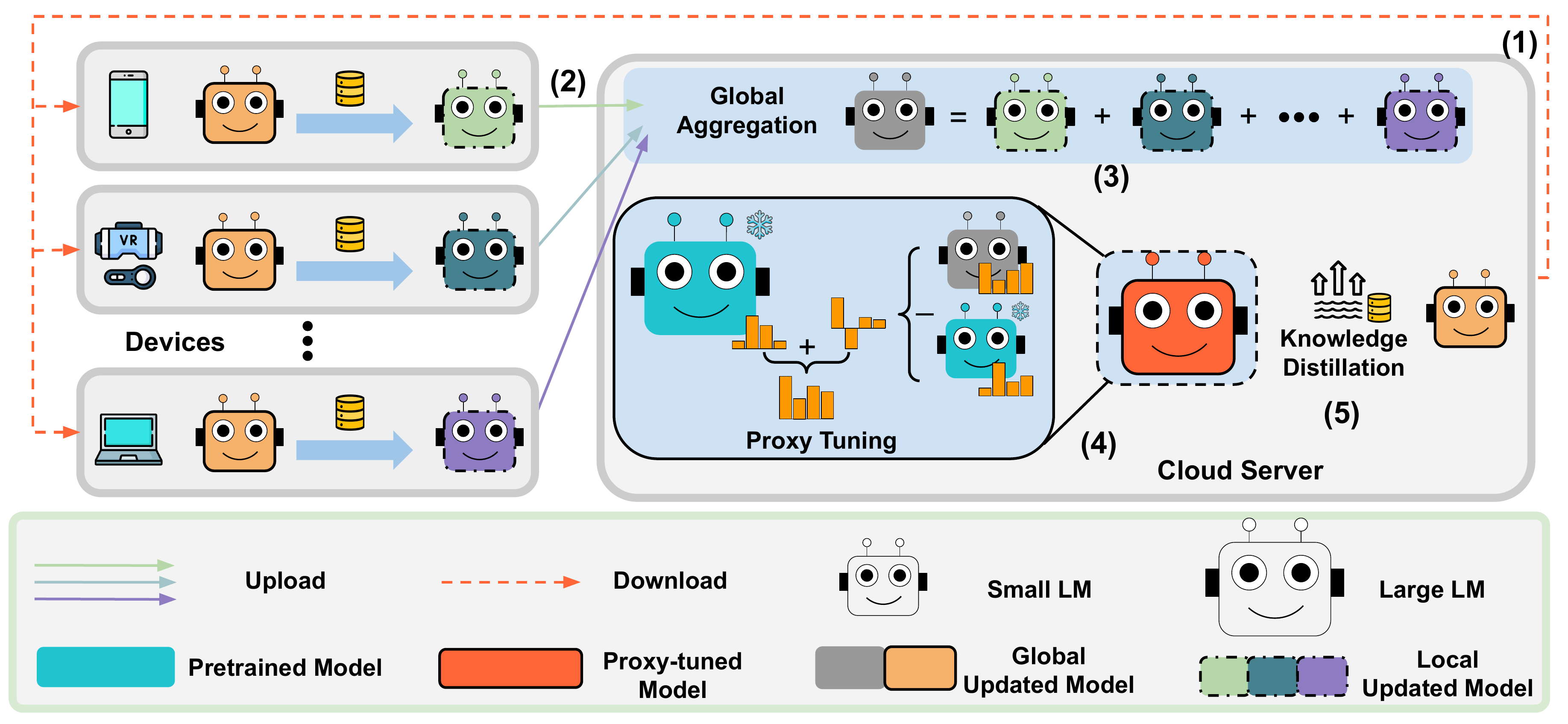}
  \caption{Overview of FedPT. In FedPT, each training round comprises the following steps: 1) The cloud server broadcasts the latest small LM to the selected devices; 2) Each selected device fine-tunes the received small LM  (e.g., using LoRA) and sends it back to the cloud server; 3) The cloud server collects and aggregates the updated small LMs; 4) The cloud server constructs the large proxy-tuned LM by utilizing the difference between the predictions of the small pre-trained and fine-tuned LMs to shift the original predictions of the larger pre-trained LM in the direction of tuning; and 5) The cloud server distills knowledge from the large proxy-tuned LM into the small aggregated LM to obtain the latest small LM. Note that FedPT does not require access to the internal model weight of the large pre-trained LM.}
  % It then sends the newly updated small LM to the selected devices for the next training round. %These steps are repeated iteratively until the proxy-tuned model meets predefined performance criteria.
  \label{fig:framework}
\end{figure*}

% \begin{enumerate*}[label={\arabic*).}]
%     % \item \emph{Memory Overhead}. {\color{green}Training large models requires significant memory, often exceeding 10GB~\cite{wang2023overview,rajbhandari2020zero}. Most devices have RAM capacities of 4-8GB~\cite{blogdoiphone2024}, which is insufficient for such tasks.}
%     \item \emph{Huge memory footprint}. Even on a GPU-equipped device, local computations can take several hundred seconds per round. Consequently, a fine-tuning session may extend over several days. For example, 
%     \item \emph{High Communication overhead}. In each FL round, participating devices are required to download the latest global model and then upload their local models, which involves transferring hundreds of MBs of data.
% \end{enumerate*}

Recently, various parameter-efficient fine-tuning (PEFT) methods have been integrated into FL to overcome the aforementioned challenges \cite{zhao2023breaking, zhao2023fedprompt, che2023federated, babakniya2023slora, cai2023efficient}. These approaches assume that devices have white-box access to a large LM's parameters, focusing on updating only a small subset of parameters. In practice, however, these assumptions do not always hold due to the following reasons. 
\begin{enumerate*}[label={\arabic*)}]
    % \item We may only have black-box access to an LLM, such as GPT-4 \cite{achiam2023gpt}, which restricts us from retrieving \hl{either the final results} or \hl{the logits from the last layer}. Since we cannot access the model parameters of these large LMs, they cannot be effectively fine-tuned; 
    \item %We may not have access to the parameters of large LMs, such as GPT-4~\cite{achiam2023gpt}, making it impractical to fine-tune these models locally.
    The pre-trained LMs could be proprietary (e.g., GPT-4), and thus their model weights are private, making it impossible to directly tune these models on edge devices in FL.  %
    % \item Limited access to the parameters of large LMs, such as GPT-4~\cite{achiam2023gpt}, may make local fine-tuning impractical.
    %
    \item Even with PEFT methods, large LM fine-tuning still requires a huge memory footprint. For example, fine-tuning a LLaMA-13B model using the LoRA~\cite{hu2021lora} requires 34.8 GB VRAM. Such requirements exceed the capabilities of most resource-constrained devices in FL. 
\end{enumerate*}

To bridge this gap, we introduce \textbf{Fed}erated \textbf{P}roxy-\textbf{T}uning (FedPT), a lightweight federated fine-tuning method for large black-box LMs that requires access only to their predictions over the output vocabulary instead of their parameters. Specifically, as demonstrated in Figure~\ref{fig:framework}, devices first collaboratively tune a \emph{smaller} LM based on their private data. Then, with the small fine-tuned LM, the cloud server constructs a large proxy-tuned LM by leveraging the difference between the predictions of the small pre-trained and fine-tuned LMs \cite{liu2024tuning, mitchell2023emulator} to shift the original predictions of the larger pre-trained LM in the direction of tuning. %
%
%
%
%align the predictions of the larger pre-trained LM. 
%
% FedPT fine-tunes a smaller LM, then contrasts the predictions of the fine-tuned small model with those of its pre-tuned version to guide the tuning of the larger pre-trained model.
% This approach makes it feasible to fine-tune the large LM in a black-box setting.
%
%FedPT operates in a black-box setting, leveraging smaller LMs that are fine-tuned locally on devices and then aggregated by the cloud server. By comparing prediction discrepancies between the aggregated small models and a pre-trained large model, the cloud server constructs a proxy-tuned large model. Knowledge from this proxy-tuned large LM is further distilled back into the aggregated small LM for further training, allowing us to fine-tune the large model without direct access to its internal parameters.
%
%FedPT operates in a black-box setting, only requiring access to prediction distributions over the output vocabulary of large LM. This approach allows us to fine-tune the large LM without accessing its internal parameters. 
After that, the server leverages knowledge distillation to transfer the knowledge from the large proxy-tuned LM to the small aggregated LM to obtain an updated small LM for further training. These steps are repeated iteratively until the large proxy-tuned LM meets predefined performance criteria. %
As will be shown later, even without access to the model weights of large LMs, FedPT can achieve comparable performance to directly federated fine-tuning of large LMs by leveraging the prediction distributions over the output vocabulary. Moreover, each device only needs to tune a \emph{smaller} LM, which can effectively reduce the computation, communication, and memory overhead in FL.

%\hl{Do we need to cite liu et al. mitchell et al? I think we need to emphasize the black box in this paragraph to response to the first challenge in the last praragraph. These steps are repeated iteratively until the proxy-tuned model meets predefined performance criteria.}

In summary, this paper makes the following main contributions: 
% \begin{enumerate*}[label={\arabic*).}]
%      \item We propose a novel federated fine-tuning framework, called FedPT, that enables resource-constrained devices to fine-tune a large LM collaboratively without sharing their private training data. 

%      \item We design a new federated fine-tuning strategy to achieve the same end as directly fine-tuning a larger LM at the cloud server by tuning only smaller LMs at devices in a federated way and applying proxy tuning and knowledge distillation to transfer the knowledge from the smaller LMs to the larger LM.   
     
%  %    knowledge distillation to accelerate the performance of the large proxy-tuned LM.
      
%      \item We conduct extensive experiments on common benchmarks to evaluate the proposed framework. Experimental results demonstrate that FedPT can significantly reduce the computation and communication costs compared with directly fine-tuning large LMs while maintaining a similar model performance.  
%  \end{enumerate*}
%
\begin{enumerate}[label={\arabic*)}]
     \item We propose FedPT, a novel FL framework, that enables resource-constrained devices to fine-tune a large black-box LM collaboratively without sharing their private training data. 

     % \item We design a new federated fine-tuning strategy to achieve the same end as directly fine-tuning a larger LM at the server by tuning only smaller LMs at devices and applying proxy tuning and knowledge distillation to transfer the knowledge from the smaller LM to the larger LMs.   

      \item We design a new federated fine-tuning strategy to achieve the same end as directly fine-tuning a larger LM at the server by tuning only smaller LMs at devices and applying proxy tuning and knowledge distillation to exchange the knowledge between the smaller LM and the larger proxy-tuned LM.   

      % \item We design a new federated proxy-tuning strategy to achieve the same end as directly fine-tuning a larger LM at the server by tuning only smaller LMs at devices and applying proxy tuning and knowledge distillation to transfer the knowledge from the smaller LM to the larger LMs. 
     
 %    knowledge distillation to accelerate the performance of the large proxy-tuned LM.
      
     \item We conduct extensive experiments on common benchmarks to evaluate the proposed framework. Experimental results demonstrate that FedPT can significantly reduce the computation, communication, and memory costs compared with directly fine-tuning large LMs while maintaining a similar model performance.  
 \end{enumerate}

\section{Problem Definition}
\label{sec:problem}

Consider an FL system for fine-tuning the large LM, which consists of a cloud server and a set of $N$ devices $\mathcal{S}$ (e.g., smartphones and IoT devices). Each device $n\in \mathcal{S}$ has a local private dataset $\mathcal{D}_n$, and the aggregated dataset of all devices is denoted as $\mathcal{D}:= \bigcup_{n \in \mathcal{S}}\mathcal{D}_n$. For $n \neq n^\prime \in \mathcal{S}$, the data distributions of $\mathcal{D}_{n}$ and $\mathcal{D}_{n^\prime}$ could be different. The goal of the FL system is to find a global large LM $\bm{\theta}_{{l}} \in \mathbb{R}^d$ with $d$ in the scale of billions that solves the following optimization problem:
\begin{equation}\label{eq: original_fl}
\min_{\bm{\theta}_{{l}} \in \mathbb{R}^d} f(\bm{\theta}_{{l}}):= \sum_{n \in \mathcal{S}} \frac{|\mathcal{D}_{n}|}{|\mathcal{D}|} f_n(\bm{\theta}_{{l}}),
\end{equation}
where $f(\bm{\theta}_{{l}})$ represents the global loss, $f_n(\bm{\theta}_{{l}}) = \mathbb{E}_{z \in \mathcal{D}_n}[\mathcal{L}(\bm{\theta}_{{l}};z)]$ is the local loss function of device $n$, $z$ denotes a datapoint sampled from $\mathcal{D}_n$, and $\mathcal{L}(\bm{\theta}_{{l}};z)$ represents the loss of model $\bm{\theta}_{{l}}$ on datapoint $z$. 

To solve the optimization problem~\eqref{eq: original_fl}, FedIT~\cite{zhang2024towards} has been proposed, which integrates FedAvg~\cite{mcmahan2017communication} with LoRA to alleviate the communication and memory overhead. However, it still faces the following challenges: (1) In practice, the white-box access to large LM $\bm{\theta}_{{l}}$ is not always available, making direct federated fine-tuning of the large LM infeasible.  
% \hl{We may only obtain the predictive distributions over the output vocabulary from a black-box large LM [Yu]} (e.g., GPT-4 \cite{achiam2023gpt}). \red{(contradictory to what we wrote in the introduction before.)} 
(2)  Even with LoRA, fine-tuning the large LM continues to demand substantial memory and computational resources due to the vast size of the large LM. 

% To address the above challenges, inspired by the concept of proxy-tuning \cite{li2022contrastive, liu2024tuning, mitchell2023emulator}, this paper develops a new FL scheme that employs federated fine-tuning of a small LM to construct a large proxy-tuned LM for closely approximating the performance of the targeted large fine-tuned LM. 

\begin{algorithm}[t]
\caption{Proposed FedPT Algorithm} 
\label{alg:1}
\textbf{Input:} Small pre-trained LM $\bm{\theta}_{{s}}^0$, large pre-trained LM $\bm{\theta}_{{l}}^0$, number of selected devices $K$, local training epochs $E$, number of training rounds $T$, public dataset $\mathcal{D}_{\text{kd}}$.\\
\textbf{Output:} Proxy-tuned model $\Tilde{\bm{\theta}}_{{l}}^T$.
\begin{algorithmic}[1]
% \STATE
% \addtocounter{ALG@line}{-1}
% \textbf{Server Executes}:\\
% \Serverexe
% \STATE  \textbf{Server Executes}:\\
%\STATE Initialize $\bm{\theta}_{{s}}^0\leftarrow\bm{\theta}_{{s}}$
\FOR{round $t=0, 1, 2 \dots, T-1$}
\STATE Server randomly selects a subset $\mathcal{S}_t$ of $K$ devices \label{alg:1:select}
\STATE Server sends $\bm{\theta}_{{s}}^t$ to all selected devices \label{alg:1:send}
\FOR{each device $k \in \mathcal{S}_t \textbf{ in parallel}$} \label{alg:1:local_begin}
% \STATE \hl{$\bm{\theta}_{{s}, i}^{t} \leftarrow$ \textbf{ClientUpdate}($\bm{\theta}_{{s}}^t$)}\label{alg:1:clientUpdate}
\STATE $\bm{\theta}_{{s}, k}^{t,E} \leftarrow$  Update $\bm{\theta}_{{s}}^t$ with $E$ epochs on $\mathcal{D}_{k}$
\STATE Send $\bm{\theta}_{{s}, k}^{t,E}$ to the server
% \textbf{LocalUpdate}($\bm{\theta}_{{s}}^t,k$)\label{alg:1:clientUpdate}
\ENDFOR \label{alg:1:local_end}
\STATE Server receives and aggregates the small averaged LM $\Bar{\bm{\theta}}_{{s}}^{t+1} \leftarrow  \sum_{k\in\mathcal{S}_t}\frac{|\mathcal{D}_k|}{\sum_{k^{\prime} \in \mathcal{S}_t } |\mathcal{D}_{k^{\prime}}|} \bm{\theta}_{{s}, k}^{t, E}$
\label{alg:1:aggregation}
\STATE Server constructs the large proxy-tuned LM $\Tilde{\bm{\theta}}_{{l}}^{t+1}$ using $\Bar{\bm{\theta}}_{{s}}^{t+1}$, $\bm{\theta}_{{s}}^0$, and $\bm{\theta}_{{l}}^0$ through Equation \eqref{eq: proxy-tuning} \label{alg:1:proxy}
% \FOR{ $h = 0,1,2,\dots,H-1$}
\STATE Server obtains $\bm{\theta}_{{s}}^{t+1}$ by distilling the knowledge from $\Tilde{\bm{\theta}}_{{l}}^{t+1}$ to $\Bar{\bm{\theta}}_{{s}}^{t+1}$ through solving Equation \eqref{eq: distillation} on $\mathcal{D}_{\text{kd}}$ \label{alg:1:kd}
% \ENDFOR
%
% \STATE Obtain the updated small model $\bm{\theta}_{{s}}^{t+1}$ 
%  according to Equation \ref{eq: distillation}
\ENDFOR
% \STATE
% \STATE \hl{\textbf{ClientUpdate}($\bm{\theta}_{{s}}^{t}$):}
% \STATE
% \[\]
% \Deviceexe
% \STATE  \textbf{ClientUpdate}($\bm{\theta}_{{s}}^{t},i$):
%\STATE Initialize $\bm{\theta}_{{s}, i}^{t,0} \leftarrow \bm{\theta}_{{s}}^{t}$
% \STATE Let $\bm{\theta}_{{s}, i}^{t} = \bm{\theta}_{{s}}^{t}$.
% \STATE  $\bm{\theta}_{{s}, k}^{t, 0} \leftarrow \bm{\theta}_{{s}}^{t}$
% \FOR{$e=0, 1, 2 \dots, E-1$}
% \STATE Compute a mini-batch stochastic gradient 
% \STATE $\bm{\theta}_{{s}, k}^{t, e+1} \leftarrow \bm{\theta}_{{s}, k}^{t, e} -\eta \nabla \mathcal{L}(\bm{\theta}_{{s}, k}^{t, e}; \mathcal{D}_k)$
% \ENDFOR
% \STATE Send the updated local model parameter $\bm{\theta}_{{s}, k}^{t, E}$ to the server
% \STATE

\end{algorithmic}
\end{algorithm}

% \section{FedPT: Federated Proxy Tuning}
\section{Methodology}
\subsection{Overview of FedPT}
% As mentioned in Section~\ref{sec:problem}, directly federated fine-tuning of large LM $\bm{\theta}_{{l}}$ is impractical. 
In this section, we introduce a new approach termed FedPT to address the aforementioned challenges. Instead of directly federated fine-tuning the large LM $\bm{\theta}_{{l}}$ to solve the optimization problem \eqref{eq: original_fl}, the goal of FedPT is to construct a large proxy-tuned LM $\Tilde{\bm{\theta}}_{{l}}$ that has similar performance as directly fine-tuning large LM $\bm{\theta}_{{l}}$ in each training round. Specifically, the large proxy-tuned LM $\Tilde{\bm{\theta}}_{{l}}$ can be decomposed into three sub-models: a small pre-trained LM $\bm{\theta}_{s}^0$, a small fine-tuned LM $\bm{\theta}_{s}$, and a large pre-trained LM $\bm{\theta}_{l}^0$. Here, the small LMs $\bm{\theta}_{s}^0, \bm{\theta}_{s} \in \mathbb{R}^{d_0}$ share the same vocabulary as the large pre-trained LM $\bm{\theta}_{l}^0 \in \mathbb{R}^{d}$ while $d \gg d_0$. Note that only the small LM $\bm{\theta}_{s}$ needs to be fine-tuned in FedPT. Essentially, FedPT aims to solve the following surrogate objective function:
\begin{equation}
    \min_{\bm{\theta}_{{s}} \in \mathbb{R}^{d_0}} f(\bm{\theta}_{{s}}):=  \sum_{n \in \mathcal{S}} \frac{|\mathcal{D}_{n}|}{|\mathcal{D}|}f_n(\bm{\theta}_{{s}}) +  \text{KL}(p(\Tilde{\bm{\theta}}_{{l}}), p(\bm{\theta}_{{s}})).
\end{equation}
The first term is the weighted averaged local training loss of small LM $\bm{\theta}_{{s}}$. The second term is the Kullback–Leibler (KL) divergence between the predicted probability distribution of the small LM $p(\bm{\theta}_{{s}})$ and large proxy-tuned LM $p(\Tilde{\bm{\theta}}_{{l}})$.

% Since the large fine-tuned LM $\bm{\theta}^{\ast}_{l}$ may is unknown in practice

% It is worth noting that large fine-tuned LM $\bm{\theta}^{\ast}_{l}$ is unknown in practice, and we will leverage \hl{the predictive distributions over the output vocabulary of proxy-tuned model [Yu]} $p(\Tilde{\bm{\theta}}_{{l}})$ to approximate $p(\bm{\theta}^{\ast}_{l})$. 

Algorithm \ref{alg:1} provides the pseudo-code of FedPT. The total training process comprises $T$ training rounds, with each round consisting of the following steps. At the beginning of $t$-th round, the cloud server randomly selects a subset $\mathcal{S}_t$ of $K$ devices and broadcasts the latest small LM $\bm{\theta}_{{s}}^{t}$ to the selected devices (Lines \ref{alg:1:select}-\ref{alg:1:send}). 
Then, each selected device $k \in \mathcal{S}_t$ performs $E$ epochs of local updates and sends the updated local model $\bm{\theta}_{{s},k}^{t,E}$ back to the cloud server (Lines \ref{alg:1:local_begin}-\ref{alg:1:local_end}). 
Next, the cloud server aggregates the local models to obtain the small averaged LM $\Bar{\bm{\theta}}_{{s}}^{t+1}$ (Line \ref{alg:1:aggregation}). 
After that, the cloud server constructs the large proxy-tuned LM $\Tilde{\bm{\theta}}_{{l}}^{t+1}$ based on the small averaged LM $\Bar{\bm{\theta}}_{{s}}^{t+1}$, small pre-trained LM $\bm{\theta}_{{s}}^0$, and large pre-trained LM $\bm{\theta}_{{l}}^0$ (Line \ref{alg:1:proxy}). The details of construction are introduced in Section \ref{sec:proxy-tuning}.
Finally, the cloud server obtains the latest small LM $\bm{\theta}_{{s}}^{t+1}$ by distilling the knowledge from $\Tilde{\bm{\theta}}_{{l}}^{t+1}$ to $\Bar{\bm{\theta}}_{{s}}^{t+1}$ (Line \ref{alg:1:kd}). The details of knowledge distillation are explained in Section \ref{sec:kd}.

It is worth noting that FedPT inherits the privacy benefits of classic FL schemes by keeping the raw data on devices and sharing only model parameters. Additionally, FedPT is compatible not only with existing PEFT methods for large LMs, such as LoRA \cite{hu2021lora}, LoHa~\cite{yeh2024navigating}, and adapter \cite{houlsby2019parameter}, but also with existing privacy-preserving techniques in FL, including secure aggregation and differential privacy.

\subsection{Proxy Tuning}\label{sec:proxy-tuning}

To deal with the huge computation, communication, and memory overheads and white-box model access requirement for the direct fine-tuning of LLMs in FL, we draw inspiration from proxy-tuning \cite{li2022contrastive, liu2024tuning, mitchell2023emulator}, which utilizes smaller LMs as proxies to guide the generation of larger LMs. For simplicity of notation, we will omit the superscript $t$ without causing ambiguity in the following. At each training round, we fine-tune a small LM $\bm{\theta}_{{s}}$, which shares the same vocabulary with the large pre-trained LM $\bm{\theta}_{{l}}^0$. Subsequently, we add a logit offset, which is defined as the difference between logits from the small fine-tuned LM $\bar{\bm{\theta}}_{{s}}$ and the pre-trained LM $\bm{\theta}_{{s}}^0$, to every token of the large pre-trained model $\bm{\theta}_{{l}}^0$ for guiding the prediction of the next word. Formally speaking, denote the input and generated sequence as $\mathbf{x} \in \mathcal{X}$ and $\mathbf{y} \in \mathcal{Y}$, respectively. Let $y_j$ be the $j$-th token in $\mathbf{y}$ and $\mathbf{y}_{<j}$ denote the sequence prefix from the beginning to the $(j-1)$-th token. Thus, the sequence-level distribution can be written as $p(\mathbf{x} | \mathbf{y}) = \prod_{j=1}p(y_j | \mathbf{x}, \mathbf{y}_{<j})$.  The probability distribution of the next word prediction from the large proxy-tuned LM $\Tilde{\bm{\theta}}_{{l}}$ can be written as 
\begin{align}
\label{eq: proxy-tuning}
    p(y_j |\mathbf{x}, \mathbf{y}_{<j}; \Tilde{\bm{\theta}}_{{l}}) := \text{softmax}\left[g(y_j |\mathbf{x}, \mathbf{y}_{<j}; \bm{\theta}_{{l}}^0) \notag \right.\\
    \qquad \left.+\alpha \left( g(y_j |\mathbf{x}, \mathbf{y}_{<j}; \Bar{\bm{\theta}}_{{s}}) - g(y_j |\mathbf{x}, \mathbf{y}_{<j}; {\bm{\theta}}_{{s}}^0) \right)\right],
\end{align}
where $g(\cdot)$ represents the logit function of the last layer of LM, and $\alpha$ is a hyperparameter that controls the amount of modification to output distribution of the large pre-trained LM. A smaller value of $\alpha$ results in predictions that closely resemble those of the large pre-trained LM, whereas a larger $\alpha$ magnifies the contrast between the small fine-tuned LM and small pre-trained LM. %
%$\alpha$ can be determined through a hyperparameter search process. %

Note that from Equation \eqref{eq: proxy-tuning}, we can obtain the following in the probability space:
\begin{equation}
    p(y_j |\mathbf{x}, \mathbf{y}_{<j}; \Tilde{\bm{\theta}}_{{l}}) \propto  g(y_j |\mathbf{x}, \mathbf{y}_{<j}; \bm{\theta}_{{l}}^0) \left(\frac{g(y_j |\mathbf{x}, \mathbf{y}_{<j}; \Bar{\bm{\theta}}_{{s}})}{g(y_j |\mathbf{x}, \mathbf{y}_{<j}; {\bm{\theta}}_{{s}}^0)}\right)^{\alpha}.
\end{equation}
From the above equation, it is evident that the small fine-tuned LM $\bar{\bm{\theta}}_{{s}}$ plays a crucial role in guiding the large proxy-tuned LM $\tilde{\bm{\theta}}_{{l}}$. Enhancing the fine-tuned $\bar{\bm{\theta}}_{{s}}$ results in a more sophisticated large proxy-tuned LM. Moreover, as detailed in Appendices~\ref{appendix: Background of Fine-Tuning LLM} and \ref{appendix: Proxy Tuning LLM}, the disparity between the large proxy-tuned LM and the directly fine-tuned large LM gradually diminishes with improvements in the small fine-tuned LM $\bar{\bm{\theta}}_{{s}}$.

\subsection{Knowledge Distillation}
\label{sec:kd}
% Since the fine-tuning datasets are significantly smaller and less diverse than web-scale pre-training datasets, there is always a risk that the fine-tuned model might ``catastrophically forget'' \cite{mccloskey1989catastrophic} the useful general knowledge learned from the pre-training stage. 
According to the previous works~\cite{li2022contrastive, liu2024tuning, mitchell2023emulator}, the proxy-tuned large LMs yield better performance compared with directly fine-tuned small LMs. Moreover, for knowledge-intensive tasks, proxy-tuning sometimes even surpasses the performance of directly fine-tuning the large LM, as it may preserve more learned knowledge than directly updating the model parameters of large LM \cite{liu2024tuning}.
% We further boost the performance of small LMs in FedPT by the proxy-tuned large LMs by knowledge distillation. 
% To tackle this issue, 
Therefore, we further leverage the knowledge distillation \cite{sanh2019distilbert, muhamed2021ctr, song2020lightpaff} to transfer the general knowledge from the teacher model (i.e., large proxy-tuned LM $\Tilde{\bm{\theta}}_{{l}}$) to the student model (i.e., small LM $\bm{\theta}_{{s}}$) in each training round. The objective of knowledge distillation is defined as follows:
\begin{multline}
\label{eq: distillation}
    \bm{\theta}_s = \argmin_{\bm{\theta}_s\in \mathbb{R}^{d_0}}\mathbb{E}_{(\mathbf{x,y})\sim \mathcal{D}_\text{kd}}\big[(1-\lambda)\text{MLE}(\mathbf{x,y}; \bm{\theta}_s)   \\
    +\lambda\text{KL}(p(\mathbf{x};\Tilde{\bm{\theta}}_{{l}}), p(\mathbf{x};\bm{\theta}_{{s}}))\big],
\end{multline}
where $(\mathbf{x,y})$ is a datapoint sampled from a small public dataset $\mathcal{D}_\text{kd}$ in the cloud server, the first term $\text{MLE}(\mathbf{x,y}; \bm{\theta})$ represents the maximum likelihood estimation of the student model $\bm{\theta}_s$, and the second term $\text{KL}(p(\mathbf{x};\Tilde{\bm{\theta}}_{{l}}), p(\mathbf{x};\bm{\theta}_{{s}}))$ denotes the KL divergence between the predicted probability distribution of the teacher model $\Tilde{\bm{\theta}}_{{l}}$ and that of the student model $\bm{\theta}_{{s}}$ on the same data sample. Here, $\lambda$ is a hyperparameter used for balancing the two loss terms (see details in Appendix \ref{append:kd}). 

%%%%%%%%%%%%%%%%%%%%%%%%%%%%%%%
\section{Experiments}
%%%%%%%%%%%%%%%%%%%%%%%%%%%%%%%
%----------------------
\subsection{Experimental Setup
}
%----------------------
We consider 10 devices in the experiments and experimentally validate the instruction-following task \cite{ouyang2022training} as a conditional text-generation task where models generate responses based onthe given instructions. Other fine-tuning tasks, such as code generation~\cite{roziere2023code,lai2023ds}, can also be applied using our approach. 

\paragraph{Models.} Our experiments utilize two distinct model families: GPT-2~\cite{radford2019language} and LLaMA~\cite{touvron2023llama}, each available in various sizes. For GPT-2 family models, we use the GPT-2-760M model as the small LM and GPT-2-1.5B as the large LM. 
% For OPT~\cite{zhang2022opt} based models, we use OPT at 1.3B, 2.7B, and 6.7B scale as small models and OPT-13B as the large model. 
For LLaMA family models, we use LLaMA-7B as the small LM, while LLaMA-13B and LLaMA-30B serve as the large LM. 
% Further, we use the proxy-tuned LLaMA-13B to distill knowledge to LLaMA-7B and then investigate the performance of LLaMA-7B in proxy-tuning LLaMA-30B. 

\paragraph{Fine-tuning Datasets.} 
For the fine-tuning dataset $\mathcal{D}$, we compile it from the ``databricks-dolly-15K''~\cite{DatabricksBlog2023DollyV2}, which contains 15,000 pairs of human-crafted instruction-following records. Specifically, we remove samples that surpass the models' context length. Then, we randomly allocate 1,000 samples for validation and 500 for testing, thereby retaining approximately 12,500 examples dedicated to training purposes. To simulate an FL setup, similar to FedIT~\cite{zhang2024towards}, we employ two data partition strategies, pathological non-IID~\cite{mcmahan2017communication} and Dirichlet non-IID~\cite{hsu2019measuring}. We present the results on pathological non-IID distribution in the main paper, and the results on the Dirichlet distribution and further details about the data heterogeneity are provided in Appendix~\ref{append:data_hete}. We utilize the Alpaca dataset~\cite{alpaca} as the public dataset $\mathcal{D}_\text{kd}$ for knowledge distillation. 

\paragraph{Evaluation Datasets.} We evaluate the performance of our federated proxy-tuned model on the following three distinct instruction-following datasets: 
\begin{enumerate*}[label={\arabic*)}]
    \item \textbf{Dolly}: A 500-sample test set derived from the databricks-dolly-15K dataset. 
    \item \textbf{SelfInst}~\cite{wang2023self}: A user-oriented instruction-following dataset with 252 samples.
    \item \textbf{S-NI}: The \textsc{Super-NaturalInstructions}~\cite{wang-etal-2022-super} test set, which includes 9,000 samples across 119 tasks. Following~\cite{peng2023instruction,gu2023minillm}, we divided this set into three subsets based on ground truth response lengths: [0, 5], [6, 10], and [11, $+\infty$]. We use the [11, $+\infty$] subset as the test set in our paper.
\end{enumerate*}

\paragraph{Evaluation Metrics.} 

We use two metrics to evaluate the model-generated responses: 
\begin{enumerate*}[label={\arabic*)}]
    \item \textbf{Rouge-L score}~\cite{lin2004rouge}: The Rouge-L score is used to assess the recall and relevance of text generated by a model by measuring the longest common subsequence of words compared to a reference text. Previous works~\cite{wang2022super,gu2023minillm} have indicated that Rouge-L is appropriate for large-scale evaluation of instruction-following tasks.
    \item \textbf{GPT-4 feedback}: We employ GPT-4-Turbo as a judge to evaluate model-generated responses from multiple perspectives, such as helpfulness, relevance, accuracy, and level of detail of their responses. The details are given in Appendix~\ref{append:gpt4score}.  
\end{enumerate*}

\paragraph{Baselines.} 
We consider three baselines in our main experiments:
\begin{enumerate*}[label={\arabic*)}]
    \item {\textbf{Base}} directly uses the base pre-trained large model on the server side.
    \item {\textbf {FedAvg}} fine-tunes the small or large LM by the FedAvg algorithm~\cite{mcmahan2017communication}. This baseline is consistent with the recent work FedIT~\cite{zhang2024towards} that focuses on instruction-following tasks. 
    \item \textbf{FedAvg+PT} follows the same procedure as FedAvg to fine-tune a small LM. During text generation, it utilizes the large proxy-tuned LM, which incorporates the small fine-tuned model, a small pre-trained model, and a large pre-trained model to generate responses. 
\end{enumerate*}

\begin{table*}[t]
  \centering
  \begin{tabular}{@{}clcccccc@{}}
\toprule
 \multirow{2}{*}{Model}& \multirow{2}{*}{Method} & \multicolumn{3}{c}{Dataset} & \multirow{2}{*}{\makecell{Model \\ Size  }} & \multirow{2}{*}{VRAM} & \multirow{2}{*}{\makecell{Comm. \\ Cost}}  \\ \cmidrule(lr){3-5}
                                             & & Dolly & SelfInst & S-NI  &     \\ 
                       % \midrule
                       % \midrule
    % Model  & Method & Dolly & SelfInst  & S-NI &\begin{tabular}{@{}c@{}}Device  Params\end{tabular}\\
     % Model  & Method & Dolly & SelfInst  & S-NI &\begin{tabular}{@{}c@{}}Device  Params\end{tabular}\\
    \midrule
     \multirow{5}{*}{\rotatebox[origin=c]{90}{LLaMA}}& Base (13B) & 9.7\textsubscript{$\pm$.2} & 7.3\textsubscript{$\pm$.5} & 8.8\textsubscript{$\pm$.1} & N/A & N/A& N/A\\
    &FedAvg (13B) & {24.5}\textsubscript{$\pm$.3} & {19.0}\textsubscript{$\pm$.8} & {29.9}\textsubscript{$\pm$.5} & 13B & 34.8GB & 2.6GB\\
    \cmidrule{2-8}
    % \cmidrule(r){6-8}
    &FedAvg (7B) & 23.3\textsubscript{$\pm$.6} & 17.6\textsubscript{$\pm$.6} & 25.9\textsubscript{$\pm$.2} & \multirow{3}{*}{7B} & \multirow{3}{*}{19.5GB}& \multirow{3}{*}{1.6GB }\\   
   & FedAvg+PT (7B-13B) & 23.5\textsubscript{$\pm$.7} & 18.9\textsubscript{$\pm$.3} & 26.7\textsubscript{$\pm$.4} &  & &\\
  &  FedPT (7B-13B) & \textbf{23.8}\textsubscript{$\pm$.4} & \textbf{19.1}\textsubscript{$\pm$.7} & \textbf{28.7}\textsubscript{$\pm$.3}  &  & &\\
  \midrule                       
   \multirow{5}{*}{\rotatebox[origin=c]{90}{GPT-2}} & Base (1.5B) & 7.2\textsubscript{$\pm$.1} & 5.5\textsubscript{$\pm$.3} & 5.8\textsubscript{$\pm$.1} & N/A &N/A &N/A\\
    
   & FedAvg (1.5B) & 19.2\textsubscript{$\pm$.4} & 11.7\textsubscript{$\pm$.7} & 22.1\textsubscript{$\pm$.4} & 1.5B  & 9.5GB &474MB\\
   \cmidrule{2-8}
    % \cmidrule(r){6-8}
  &  FedAvg (760M) & 17.8\textsubscript{$\pm$.5} & 10.4\textsubscript{$\pm$.3} & 18.4\textsubscript{$\pm$.3} & \multirow{3}{*}{760M} & \multirow{3}{*}{6.1GB} & \multirow{3}{*}{286MB } \\
  &  FedAvg+PT (760M-1.5B)& 18.6\textsubscript{$\pm$.4} & 10.9\textsubscript{$\pm$.5} & 21.4\textsubscript{$\pm$.3} &  & &\\
  &  FedPT (760M-1.5B)& \textbf{18.9}\textsubscript{$\pm$.5} & \textbf{11.0}\textsubscript{$\pm$.4} & \textbf{21.6}\textsubscript{$\pm$.2} &   & &\\
    \bottomrule
  \end{tabular}
    \caption{Evaluation results. We report the average and standard deviation of Rouge-L scores across 5 random seeds. Higher values indicate better performance. \textbf{Model Size} indicates the size of the model deployed on each device. \textbf{VRAM} indicates the memory required to train one sample. \textbf{Comm. Cost} represents the total communication overhead among all devices across 20 rounds.} \label{tab:rouge}
\end{table*}
%%%%%%%%%%%%%%%%%%%%%%%%%%%%%%%%%%%%%%%%%%%%%

\begin{figure*}[t]
\centering
    \subfloat[Dolly]{\centering{\includegraphics[width=0.23\textwidth]{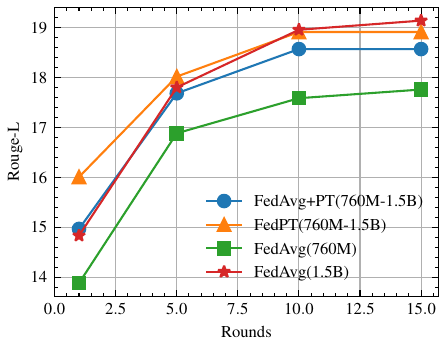} } }
    \hspace{1cm}
    \subfloat[SelfInst]{\centering{\includegraphics[width=0.23\textwidth]{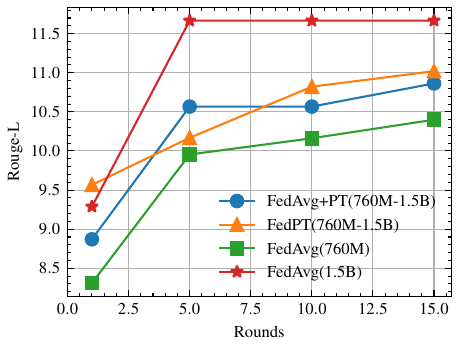} } }
    \hspace{1cm}
     \subfloat[S-NI]{\centering{\includegraphics[width=0.23\textwidth]{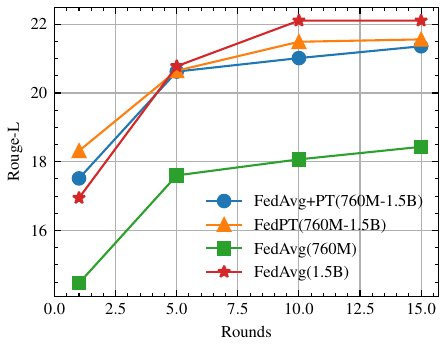} } }\\
     \subfloat[Dolly]{\centering{\includegraphics[width=0.23\textwidth]{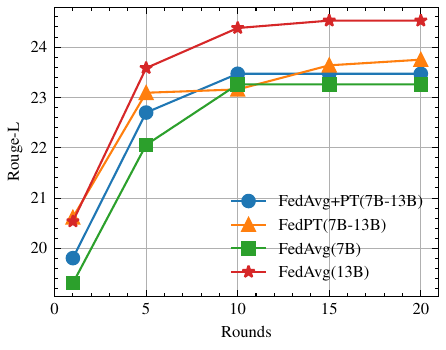} } }
     \hspace{1cm}
    \subfloat[SelfInst]{\centering{\includegraphics[width=0.23\textwidth]{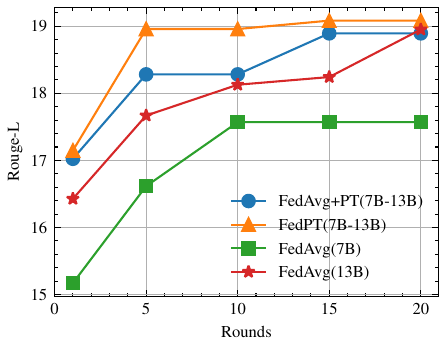} } }
    \hspace{1cm}
     \subfloat[S-NI]{\centering{\includegraphics[width=0.23\textwidth]{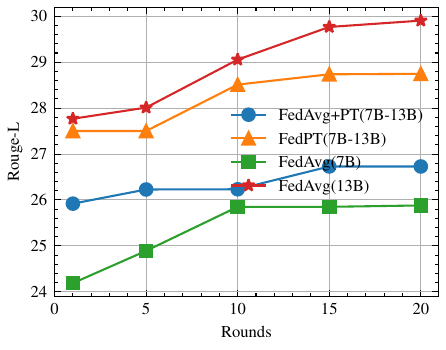} } }
    \caption{ Evaluation results of FedPT and baselines on LLaMA (a, b, c) and GPT-2 (d, e, f) models across different rounds for Dolly, SelfInst, and S-NI datasets. Higher Rouge-L scores indicate better performance.}    \label{fig:rouge}%
\end{figure*}

\begin{figure*}[ht]
\centering
    \subfloat[Dolly]{\centering{\includegraphics[width=0.23\textwidth]{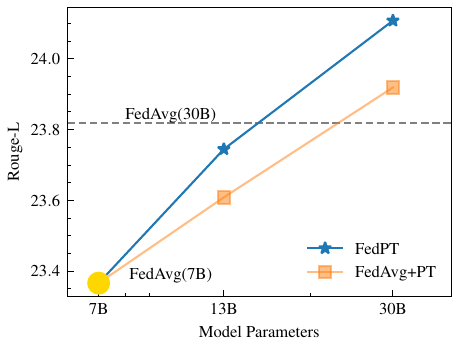} } }
    \hspace{1cm}
    \subfloat[SelfInst]{\centering{\includegraphics[width=0.23\textwidth]{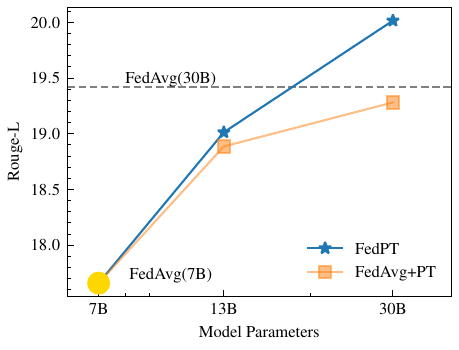} } }
    \hspace{1cm}
    \subfloat[S-NI]{\centering{\includegraphics[width=0.23\textwidth]{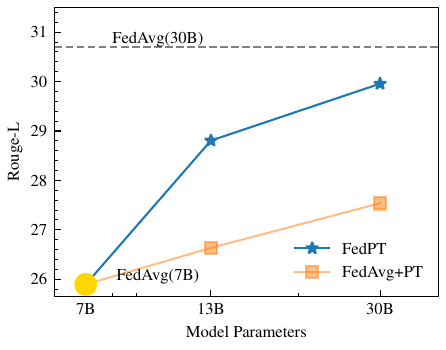} } }\\
    \caption{The scaling law of proxy-tuned models in the LLaMA family. FedAvg (7B) and FedAvg (30B) are directly fine-tuned models by FedAvg. At the 13B scale, we report the performance of FedPT (7B-13B) and FedAvg+PT (7B-13B). At the 30B scale, we use the fine-tuned 7B model from FedPT (7B-13B) to proxy-tune the 30B model for FedPT, and the 7B model from FedAvg (7B) to proxy-tune the 30B model for FedAvg+PT.}
    \label{fig:scaling}%
\end{figure*}

%-------------------------------
\begin{figure*}[t]
\centering
    % \subfloat[Dolly]{\centering{\includegraphics[width=0.3\textwidth]{./Figures/fedpt/gpt2-alpha/dolly_Rouge-L.pdf} } }
    % \subfloat[SelfInst]{\centering{\includegraphics[width=0.3\textwidth]{./Figures/fedpt/gpt2-alpha/self_inst_Rouge-L.pdf} } }
    % \subfloat[S-NI]{\centering{\includegraphics[width=0.3\textwidth]{./Figures/fedpt/gpt2-alpha/sinst_Rouge-L.pdf} } }\\
    \subfloat[Dolly]{\centering{\includegraphics[width=0.23\textwidth]{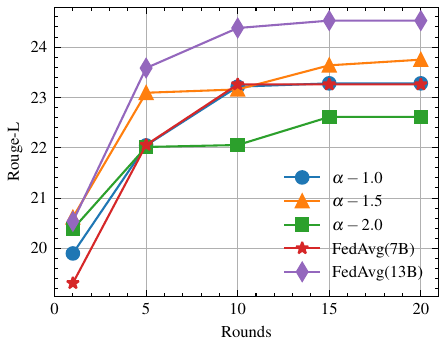} } }
    \hspace{1cm}
    \subfloat[SelfInst]{\centering{\includegraphics[width=0.23\textwidth]{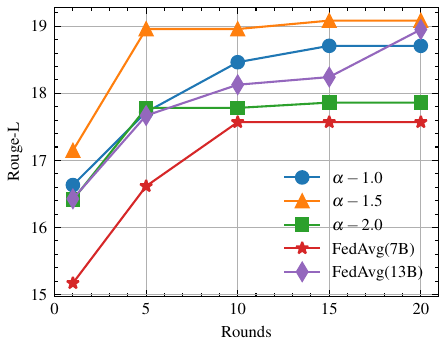} } }
    \hspace{1cm}
    \subfloat[S-NI]{\centering{\includegraphics[width=0.23\textwidth]{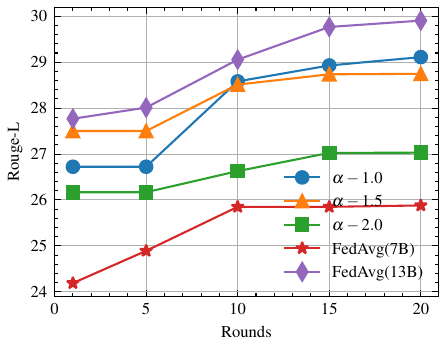} } }
    \caption{Performance comparison of different $\alpha$ for FedPT on LLaMA across different rounds for Dolly, SelfInst, and S-NI datasets. Higher Rouge-L scores indicate better performance.}
    \label{fig:effect_alpha_llama}%
\end{figure*}

\paragraph{Hyperparameters.} 
% For all experiments, we set the local training batch size to 64 and the total number of communication rounds to 20. We use the Prodigy optimizer~\cite{mishchenko2024prodigy}, an adaptive parameter-free learner, with a weight decay of 0.02. We use safeguard warmup without bias correction for all experiments. The initial learning rate is set to 1 and decayed with cosine strategy~\cite{loshchilov2016sgdr} at each communication round. For knowledge distillation of FedPT, we randomly sample 128 and 512 instances from the Alpaca dataset for the experiment on GPT-2 and LLaMA, respectively. For all experiments, we load the model into VRAM in half-precision, and the checkpoints saved during training are also in half-precision. %
%
In all experiments, we use the most common PEFT technique, LoRA~\cite{hu2021lora}, for our local training (see Appendix~\ref{appendix: PEFT} for more details). Detailed parameters about LoRA can be found in Appendix~\ref{append:lora_parameters}. We fine-tune the models for 20 communication rounds using the Prodigy optimizer~\cite{mishchenko2024prodigy}, with a batch size of 64 and an initial learning rate of 1. A cosine learning rate decay strategy~\cite{loshchilov2016sgdr} is applied at each communication round, and safeguard warmup without bias correction is implemented. To save the memory footprint, all models are loaded into VRAM in half-precision mode, with checkpoints also saved in this format. For knowledge distillation in FedPT, the hyperparameter $\lambda$ is set to $0.1$. 128 and 512 instances are sampled from the Alpaca dataset for the experiment on GPT-2 and LLaMA, respectively. More details are shown in Appendix~\ref{append:hyper-para}.

\paragraph{Experiment Overview.} We conduct experiments on GPT-2 and LLaMA models, recording checkpoints at each communication round and evaluating their performances on the three test datasets. For GPT-2 models, we conduct evaluations at communication rounds 1, 5, 10, and 15, as performance plateaued after round 10. For LLaMA models, evaluations are performed at communication rounds 1, 5, 10, 15, and 20. This results in 12 checkpoints for GPT-2 evaluation and 15 checkpoints for LLaMA evaluation. We search for the optimal $\alpha$ for both FedAvg+PT and FedPT, selecting the best $\alpha$ from $\{1.0, 1.3, 1.5, 1.8, 2.0\}$ for the GPT-2 model and $\{1.0, 1.5, 2.0\}$ for the LLaMA model. Finally, we obtain 720 evaluation results for the GPT-2 model and 600 evaluation results for the LLaMA model, totaling 1,320 evaluation results across all experiment settings.

%--------------------------------------------
\subsection{Experimental Results}\label{results}
%--------------------------------------------
\begin{table}
    \centering
    \begin{tabular}{lccc}
        \toprule
        {Method} & {Dolly} &{SelfInst} & {S-NI} \\
        \midrule
        FedAvg (13B) & 65.4 & 59.5 & 61.8 \\
        \midrule
        FedAvg (7B) & 57.7 & 52.1 & 50.7 \\
        FedAvg+PT (7B-13B)  & 63.6 & 56.4 & 59.0 \\
        FedPT (7B-13B) & \textbf{65.4} & \textbf{60.3} & \textbf{61.6} \\
        \bottomrule
    \end{tabular}
        \caption{Evaluation results by GPT-4 feedback on LLaMA. Higher scores indicate better performance.}
    \label{tab:gpt4_eval_llama}
\end{table}

We first conduct a comprehensive comparison of FedPT and the baselines across Dolly, SelfInst, and S-NI datasets in terms of model size, resource consumption, and model performance. The final results are summarized in Table~\ref{tab:rouge}, and the detailed results during training are depicted in Figure~\ref{fig:rouge}. 

%Table~\ref{tab:rouge} provides a comprehensive comparison of FedPT and the baselines across Dolly, SelfInst, and S-NI datasets.
% using Rouge-L scores. This table also includes each device's associated training model size, VRAM usage, and total communication cost among 20 rounds for LLaMA and GPT-2. 
% Specifically, model size is the \hl{measurement} dimension of the deployed fine-tuning model on each device, VRAM indicates the amount of memory required for each model \hl{on each device}, and the communication cost is the total communication overhead across all training rounds. 
%
\paragraph{Model Size and Resource Costs.}  For both FedPT and FedAvg+PT, the large pre-trained model is only utilized on the server side, allowing small LMs to be deployed on each client. This setup significantly reduces memory, storage, and communication costs. As shown in Table~\ref{tab:rouge}, compared to FedAvg with LLaMA-13B, both FedPT and FedAvg+PT achieve a $44\%$ reduction in VRAM usage and a $36\%$ reduction in communications costs. Similarly, compared to FedAvg with GPT-2-1.5B, both algorithms attain a $38\%$ reduction in VRAM usage and a $40\%$ reduction in communication costs. Here we record the VRAM usage with the local training batch size of one. Note that VRAM usage may vary slightly due to different implementation details and hardware conditions. The reported VRAM consumption is not suitable for most devices but can be reduced through quantization~\cite{dettmers2024qlora} and CPU offloading~\cite{ren2021zero}. For LLaMA experiments, we employ the model parallelism with gradient accumulation to avoid VRAM overflow.

%\hl{(can be moved to the appendix if the space is not sufficient.)}

% the model size on each device remains the same as that of the small models fine-tuned through FedAvg. Specifically,  They significantly save the VRAM usage by 44\% reduction compared to FedAvg on LLaMA-13B and GPT-2-1.5B. 

% federated fine-tuning small models.thus, they can significant reduce the model size and VRAM usage, and communication cost on e

% As shown in Table~\ref{tab:rouge}, for FedAvg+PT, the large pre-trained model is utilized after the federated fine-tuning of the small model.  For example, FedAvg+PT only needs to deploy LLaMA-7B and GPT-2-760M on each device. For FedPT, in each communication round, after the global aggregation, the large pre-trained model is utilized as part of the proxy-tuned model for knowledge distillation on the server side, meaning each device only handles the small models. Both FedAvg+PT and FedPT ensure efficient resource usage, such as VRAM and communication cost, by limiting the model size on each device to small models, while leveraging the large model for enhanced performance. \hl{our reduce, reason}

\paragraph{Model Performance Comparison.} From Table~\ref{tab:rouge}, we have the following observations for the base pre-trained and directly federated fine-tuning methods. First, the base pre-trained models suffer from inferior performance across all downstream datasets. %
% For instance, Table~\ref{tab:rouge} shows that base pre-trained LLaMA-13B achieves Rouge-L scores of 9.7 on the Dolly dataset, while pre-trained GPT-2-1.5B scores 7.2 on Dolly. 
Second, directly fine-tuning the large LM through the FedAvg method can significantly improve the model performance on specific downstream tasks. 
% For example, the LLaMA-13B model's Rouge-L score improves from 9.7 to 24.5. Meanwhile, it almost achieves the best results for both LLaMA and GPT-2 models. 

Then for the methods that use proxy-tuning, we have three observations. First, proxy-tuning can achieve performance comparable to the direct fine-tuning of the large LMs in the FL setting. For instance, although FedAvg+PT and FedPT only fine-tune GPT-2-760M locally, they achieve Rouge-L scores of 18.9 and 18.6, respectively. These scores surpass the 17.8 achieved by directly fine-tuning GPT-2-1.5B using FedAvg and are nearly as high as the 19.2 achieved by fine-tuning GPT-2-1.5B with FedAvg. Second, FedPT consistently outperforms FedAvg+PT. This superior performance is attributed to the use of a proxy model in FedPT to conduct knowledge distillation during the training process, thereby enhancing the performance of the aggregated smaller model. Third, we notice that FedPT exceeds the performance of FedAvg on LLaMA-13B for the SelfInst dataset. A similar phenomenon is also observed in~\cite{liu2024tuning}. This consistency suggests that proxy-tuning large models may better preserve knowledge more effectively than direct fine-tuning, which could potentially degrade performance on knowledge-intensive tasks. This highlights the great potential of the proxy-tuning approach. 

In addition to the results on Rouge-L score, we also present the GPT-4 feedback results for LLaMA in Table~\ref{tab:gpt4_eval_llama}. Due to the poor performance of the base pre-trained large model, we have excluded its GPT-4 score in Table~\ref{tab:gpt4_eval_llama}. Additional results for GPT-2 are given in Appendix~\ref{append:gpt4score}. Note that we have the same observations for the GPT-4 evaluation results.

\paragraph{Scaling Law.}
We first investigate the performance of FedPT and FedAvg+PT when we scale up the size of a large model in Figure~\ref{fig:scaling}. Specifically, we evaluate FedPT on LLaMA-30B, reusing the fine-tuned LLaMA-7B from FedPT (7B-30B). Similarly, we evaluate FedAvg+PT on LLaMA-30B, reusing the fine-tuned LLaMA-7B from FedAvg (7B). This approach is designed to simulate a realistic scenario in which, during the training phase, only LLaMA-7B and LLaMA-13B are used. In the deployment phase, however, if more powerful models such as LLaMA-30B become available, we aim to evaluate whether the model trained with FedPT retains its advantage over FedAvg+PT. From Figure~\ref{fig:scaling}, we observe that performance improves for both FedPT and FedAvg+PT as the model size increases. The results for both methods show a clear positive correlation between model size and performance, highlighting the scaling law: larger models yield better results. Additionally, FedPT consistently outperforms FedAvg+PT, reinforcing the advantage of our proposed method as the number of model parameters increases.

%----------------------
% \paragraph{Scaling Law of proxy model}(minillm)
% Llama: Dolly dataset

%%%%%%%%%%%%%%%%%%%%%%%%%%%%%%%%%%%%
% \paragraph{Proxy-tuned LLaMA-30B}\label{subsec:llama30b}
%%%%%%%%%%%%%%%%%%%%%%%%%%%%%%%%%%%%

%%%%%%%%%%%%%%%%%%%%%%%%%%%%%%%%%%%%
\paragraph{Effect of $\bm{\alpha}$.}\label{subsec:alpha}
%%%%%%%%%%%%%%%%%%%%%%%%%%%%%%%%%%%%
We then use various $\alpha$ values in FedPT to investigate the effect of proxy-tuning weight $\alpha$. As shown in \eqref{eq: proxy-tuning}, the generated logit follows $g_{\bm{\theta}_{{l}}^0}+\alpha (g_{\Bar{\bm{\theta}}_{{s}}} - g_{{\bm{\theta}}_{{s}}^0})$. Intuitively, larger $\alpha$ magnifies the influence of the difference between the fine-tuned small model and the pre-trained small model, making the predictions more responsive to the fine-tuning adjustments. Conversely, a smaller $\alpha$ results in predictions more similar to the large pre-trained model, causing the predictions to adhere to the behavior of the original large pre-trained model closely. 

Figure~\ref{fig:effect_alpha_llama} shows the results of $\alpha\in\{1.0, 1.5, 2.0\}$ for LLaMA on the three datasets. Specifically, $\alpha=1.5$ yields the best performance for the Dolly and SelfInst dataset, while $\alpha=1.0$ performs best for the S-NI dataset.
This demonstrates the importance of carefully tuning $\alpha$ to balance the trade-off between leveraging fine-tuning adjustments and maintaining the stability of the pre-trained model's predictions. Results for GPT-2 are provided in Appendix~\ref{append:alpha_appendix}.

% %%%%%%%%%%%%%%%%%%%%%%%%%%%%%%%%%%%%
% \paragraph{Other Analyses.}
% %%%%%%%%%%%%%%%%%%%%%%%%%%%%%%%%%%%%
% We also conduct the evaluations on the tokens most influenced by proxy-tuning and generation diversity (see details in Appendices \ref{append:top_tokens} and \ref{append:generation diversity}). The fine-grained performance for each category of the three test datasets can be found in Appendix~\ref{append:category}. Last, we present the example generations in Appendix~\ref{append:example_generation}. \red{(This paragraph is not needed while the appendicies can be kept.)}

% In Figure~\ref{fig:effect_alpha_llama}, we can find that the value $\alpha$ plays a crucial role in determining the model's behavior. 

%---------------------------------
% \begin{figure}[t]
%     \subfloat[Dolly]{{\includegraphics[width=0.3\textwidth]{./Figures/fedpt/llama-alpha/dolly_Rouge-L.pdf} } }
%     \subfloat[SelfInst]{{\includegraphics[width=0.3\textwidth]{./Figures/fedpt/llama-alpha/self_inst_Rouge-L.pdf} } }
%     \subfloat[S-NI]{{\includegraphics[width=0.3\textwidth]{./Figures/fedpt/llama-alpha/sinst_Rouge-L.pdf} } }
%     \caption{Effect of $\alpha$ in evaluation for FedPT on LLaMA.}
%     \label{fig:effect_alpha_llama}%
% \end{figure}

% \paragraph{Effect of $\alpha$}(proxy-tuning)
% GPT-2: Dolly dataset: inference alpha.

% \paragraph{Effect of Knowledge Distillation}
% GPT-2: Dolly dataset

% \paragraph{Evaluation on different categories.} 
% GPT-2: Dolly dataset

%%%%%%%%%%%%%%%%%%%%%%%%%%%%%%%
\section{Related work}
%%%%%%%%%%%%%%%%%%%%%%%%%%%%%%%

%------------------------
\paragraph{Federated Fine-tuning of Large LMs.}
% \subsection{}
%------------------------
Although fine-tuned large LMs has demonstrated remarkable success across various domain-specific NLP tasks, deployment of large LMs is hindered by significant resource demand and data privacy concern. %%For example, individuals may be reluctant to share their sensitive raw data. %
Federated fine-tuning of large LMs has been proposed as a promising technique to address the privacy concern, which enables multiple devices to fine-tune the large LM without sharing their private data. However, FL environments introduce stringent resource constraints, particularly on resource-constrained edge devices. This dilemma has catalyzed a shift towards integrating PEFT methods with the FL framework \cite{zhao2023breaking, zhao2023fedprompt, che2023federated, babakniya2023slora, cai2023efficient}. For example, federated prompt tuning is introduced in \cite{zhao2023breaking, zhao2023fedprompt, che2023federated}, which only updates the soft prompt in each communication round of FL.  %%\hl{These approaches assume that devices have enough memory footprint to fine-tune the large LM model which is not practical.} % 
Recent work \cite{xu2024fwdllm} introduced a backpropagation-free FL framework for training large LMs so that it can reduce the required memory footprint effectively. However, these works are all based on the assumption of white-box access to large LMs. In contrast, FedPT only needs to fine-tune a small LM on each client while assuming black-box access to large LMs at the server, making it more appealing in practice. %Meanwhile, FedPT can achieve a similar performance as directly fine-tuning the large LM.

% PEFT has emerged as a promising technique to adapt large pre-trained models to specific downstream tasks without retraining or fine-tuning all the model parameters. This approach is particularly beneficial for limited computational and storage resources or the adaptation needs to be rapidly deployed. Recently, several innovative PEFT methods have been proposed, 

%------------------------
\paragraph{Decoding-time Tuning.}
% \subsection{}
%------------------------

Recent advancements in large LM applications have introduced a novel approach that ``tunes'' large LMs at decoding time. One such method, known as contrastive decoding~\cite{li2023contrastive}, improves text generation quality by subtracting the log probabilities of a smaller LM (called the amateur) from a larger LM (called the expert). This approach is subject to a plausibility constraint, ensuring that the generated text surpasses the quality produced by the large LM alone. Motivated by the logit-based tuning, a collaborative generation framework was proposed, merging logits from a small LM and a large LM through a learnable model to address privacy concerns~\cite{zhang2024cogenesis}. Similarly, a method that merges output probability distributions from a small LM and a large LM through a learned small network was developed in~\cite{ormazabal2023comblm}. The most recent studies~\cite{mitchell2023emulator,liu2024tuning} utilize differences in logits as significant weights to recalibrate the conditional distributions within the large LM, thereby enhancing text generation capabilities. Specifically, one study analyzed the contribution of scaling up fine-tuning or pre-training~\cite{mitchell2023emulator}, while another demonstrated the effectiveness of merging output logits from multiple LMs~\cite{liu2024tuning,liu2021dexperts}. However, all of these studies focuses on centralized training, and hence their approaches are not applicable to the FL setting considered in our work.

\section{Conclusion}
In this work, we propose a novel FL framework called FedPT, designed for efficient fine-tuning of large LMs on resource-constrained devices without compromising privacy. A key advantage of FedPT is its ability to fine-tune large LMs without requiring access to their full model parameters. The experimental results confirm that FedPT achieves performance comparable to direct federated fine-tuning of large models, while significantly reducing resource costs in terms of storage, VRAM usage, and communication overhead. 
\bibliography{reference}

\appendix
% \appendices
% \begin{appendices}
%     \section{123}\label{23}
% \end{appendices}
\section{Background}

\subsection{Background of Fine-Tuning LLM} 
\label{appendix: Background of Fine-Tuning LLM}
The general fine-tuning of the LMs process can be treated as a reinforcement learning (RL) process \cite{rafailov2024direct}. We initial a policy $\pi=\bm{\theta}_{{l}}$, and then fine-tune $\pi$ to perform the task well. We denote the scalar-valued reward function $r: \mathcal{X} \times \mathcal{Y} \rightarrow \mathbb{R}$, which represents the human preference of a response $y$ to the query $x$. The RL goal is to maximize the expected rewards:
\setcounter{equation}{5}
\begin{equation}
    \max_{\pi}\mathbb{E}_{x \sim \mathcal{D}, y \sim \pi(\cdot|x)} [r(x,y)],
\end{equation}
where $\mathcal{D}$ is a fixed distribution (or dataset) of prompts. However, directly maximizing expected rewards can lead to a distribution collapse, which reduces the fluency and diversity of samples from the LLM \cite{korbak2022rl}. In order to solve the distribution collapse problem, one effective strategy is to include preserving the distributional properties of an LLM as part of the reward function \cite{korbak2022rl, mitchell2023emulator, ziegler2019fine, liu2024decoding}. The reformulated RL goal can be written as
\begin{equation}
   \max_{\pi}\mathbb{E}_{x \sim \mathcal{D}, y \sim \pi(\cdot|x)} [r(x,y)-\beta\text{KL}(\pi(\cdot | x) || \pi_0(\cdot | x))],
\end{equation}
where $\pi_0$ is the pre-trained model. The penalty Kullback-Leibler (KL) divergence term $\beta\text{KL}(\pi(\cdot | x) || \pi_0(\cdot | x))$ can keep $\pi$ from moving too far from $\pi_0$. Here, $\beta$ controls the strength of the KL constraint to the pre-trained model. The closed-form solution \cite{ziegler2019fine, korbak2022rl, raffel2020exploring, rafailov2024direct} can be shown as 
\begin{equation}
\label{eq: pt}
    \pi^*(y|x)=\frac{1}{Z(x)}\pi_0(y|x)\exp{\left(\frac{1}{\beta}r(x,y)\right)},
\end{equation}
where $Z(x)=\sum_y \pi_0(y|x)\exp{\left(\frac{1}{\beta}r(x,y)\right)}$. According to \cite{rafailov2024direct}, we utilize the $r_\pi(y|x)=\beta \log{\frac{\pi(y|x)}{\pi_0(y|x)}}$ as the reward function which is implicitly defined by the LM $\pi$ and $\pi_0$. Then, we can reformulate Equation~\eqref{eq: pt} as
\begin{equation}
\label{eq: final_ft}
    \pi^*(y|x) = \pi_0(y|x)\exp{\left(\log{\frac{\pi(y|x)}{\pi_0(y|x)}}\right)}.
\end{equation}
From the above equation, we can observe that the base log probabilities represent the knowledge acquired during pre-training, whereas the capabilities gained through fine-tuning are reflected in the reward (e.g., the difference, calculated by subtracting the base log probabilities from the fine-tuned model log probabilities, indicates the improvement gained from fine-tuning.).

\subsection{Proxy-Tuning LLM} 
\label{appendix: Proxy Tuning LLM}
Based on the observation of Equation~\eqref{eq: final_ft}, we fine-tune a small pre-trained model $\pi_{\bm{\theta}_{{s}}}$, which shares the same vocabulary with the large pre-trained model $\pi_{\bm{\theta}_{{l}}}$. Instead of directly fine-tuning the large pre-trained model  $\pi_{\bm{\theta}_{{l}}}$, we proxy-tune LLM \cite{mitchell2023emulator}:
\begin{align}
    \label{eq: final_pt}
    \Tilde{\pi}(y|x) &= \frac{1}{\Tilde{Z}_0(x)}\pi_{\bm{\theta}_{{l}}}^0(y|x)\exp{\left(r_{\bm{\theta}_{{s}}}(x,y)\right)} \notag \\
    &\propto \pi_{\bm{\theta}_{{l}}}^0(y|x) \frac{\pi_{\bm{\theta}_{{s}}}(y|x)}{\pi_{\bm{\theta}_{{s}}}^0(y|x)},
\end{align}
where $\pi_{\bm{\theta}_{{l}}}^0$ and $\pi_{\bm{\theta}_{{s}}}^0$ are the pre-trained LLM and small LM, respectively. The reward function $r_{\bm{\theta}_{{s}}}(x,y)=\log\frac{\pi_{\bm{\theta}_{{s}}}(y|x)}{\pi_{\bm{\theta}_{{s}}}^0(y|x)}$ and $\Tilde{Z}_0(x)=\sum_y \pi_{\bm{\theta}_{{l}}}^0(y|x)\exp{\left(r_{\bm{\theta}_{{s}}}(x,y)\right)}$. Note that it is expensive to estimate the partition function $\Tilde{Z}_0(x)$ \cite{rafailov2024direct, dudik2015contextual}. In our experiments, we utilize a per-timestep approximation and rewrite Equation~\eqref{eq: final_pt} as
\begin{align}
\label{eq:final_pt_sequence}
        \Tilde{\pi}(y_j |\mathbf{x}, \mathbf{y}_{<j}) &= 
        \frac{1}{\Tilde{Z}_1(\mathbf{x}, \mathbf{y}_{<j})}\pi_{\bm{\theta}_{{l}}}^0(y_j |\mathbf{x}, \mathbf{y}_{<j})\exp{\left(r_{\bm{\theta}_{{s}}}(\mathbf{x}, \mathbf{y}_{<j})\right)} \notag \\
        &\propto \pi_{\bm{\theta}_{{l}}}^0(y_j |\mathbf{x}, \mathbf{y}_{<j}) \frac{\pi_{\bm{\theta}_{{s}}}(y_j |\mathbf{x}, \mathbf{y}_{<j})}{\pi_{\bm{\theta}_{{s}}}^0(y_j |\mathbf{x}, \mathbf{y}_{<j})},
\end{align}
where $r_{\bm{\theta}_{{s}}}(\mathbf{x}, \mathbf{y}_{<j})=\log\frac{\pi_{\bm{\theta}_{{s}}}(y_j|\mathbf{x}, \mathbf{y}_{<j})}{\pi_{\bm{\theta}_{{s}}}^0(y_j|\mathbf{x}, \mathbf{y}_{<j})}$ and $\Tilde{Z}_1(\mathbf{x}, \mathbf{y}_{<j})=\sum_y \pi_{\bm{\theta}_{{l}}}^0(y_j|\mathbf{x}, \mathbf{y}_{<j})\exp{\left(r_{\bm{\theta}_{{s}}}(\mathbf{x}, \mathbf{y}_{<j})\right)}$. Moreover, in order to better control the impact of the small fine-tuned model, we reformulate the Equation~\eqref{eq:final_pt_sequence} as 
\begin{equation}
        \Tilde{\pi}(y_j |\mathbf{x}, \mathbf{y}_{<j}) \propto \pi_{\bm{\theta}_{{l}}}^0(y_j |\mathbf{x}, \mathbf{y}_{<j}) \left( \frac{\pi_{\bm{\theta}_{{s}}}(y_j |\mathbf{x}, \mathbf{y}_{<j})}{\pi_{\bm{\theta}_{{s}}}^0(y_j |\mathbf{x}, \mathbf{y}_{<j})} \right)^\alpha,
\end{equation}
where a small value of $\alpha$ results in predictions that closely resemble those of the original LLM, whereas a larger $\alpha$ produces predictions that are more similar to those of the small fine-tuned model.

%————————————————————————————————————————————————————
\subsection{Parameter Efficiency (LoRA) in Federated Learning}
\label{appendix: PEFT}
%————————————————————————————————————————————————————
FL environments introduce stringent resource constraints, particularly on edge devices. This dilemma has catalyzed a pivot towards PEFT methods, such as LoRA~\cite{hu2021lora}, LoHa~\cite{yeh2024navigating}, P-Tuning~\cite{liu2023gpt}, Prefix Tuning~\cite{li2021prefix}, and Prompt Tuning~\cite{lester2021power}. Our approach is compatible with many PEFT methods. Here, we have chosen the most commonly used technique LoRA, which freezes the pre-trained model weights and introduces trainable low rank metrics into each layer of the transformer architecture, significantly reducing the number of trainable parameters needed for downstream tasks. Specifically, we freeze the pre-trained weight matrix $W_0\in\mathbb{R}^{d,k}$, and constrain its update by representing it using a low-rank decomposition $W_0+\Delta W=W_0+BA$, where $B\in\mathbb{R}^{d,r}$, $A\in\mathbb{R}^{r,k}$ are two trainable parameters and the rank $r\ll\min(d,k)$. Thus, for a linear layer $h=W_0x$, the modified forward pass yields:
\begin{equation}\label{eq:lora}
    h=W_0x+\Delta Wx= W_0x+BAx.
\end{equation}
We use a random Gaussian initialization for $A$ and zero for $B$, such that $\Delta W=BA=0$ at the beginning of training. Compared to fully fine-tuning, LoRA significantly reduces memory and storage usage on local devices. 

In the FL setting, deploying LoRA only requires devices to transmit the low-rank matrices $A$ and $B$ to the server, substantially reducing communication costs compared to scenarios where full model updates are sent. The server then aggregates these matrices using FedAvg as detailed in \cite{mcmahan2017communication}. Notably, LoRA does not introduce additional latency during inference, unlike full model fine-tuning, and it offers scalability by allowing adjustments to the rank $r$. 

% \begin{table}
%   \caption{Numbers of parameters (frozen\&trainable), and GPU memory cost on a single GPU.}
%   \label{table:parameters}
%   \centering
%   \begin{tabular}{l|c|c|c|c|c}
%     \toprule
%     Model &\ Orig. Param & Adapt.Param &GPU Memory \\
%     \midrule
%     GPT-2-Large & 760M & ??M & ??\\
%     GPT-2-Xlarge & 1.5B & ??M & ??\\
%     LLaMA 7B & 7B & ??M & ??\\
%     LLaMA 13B & 13B & ??M & ??\\
%     % LLaMA 13B & 13B & ??M & ?? & ?? & ??\\
%     % LLaMA 13B & 13B & ??M & ?? & ?? & ??\\
%     \bottomrule
%   \end{tabular}
% \end{table}

%————————————————————————————————————————————————————
\subsection{Knowledge Distillation in NLP Tasks}\label{append:kd}
%————————————————————————————————————————————————————

In the field of NLP, numerous studies have implemented knowledge distillation for text classification and text generation tasks. For text classification tasks, these works enhance the performance of the student model by aligning it with the teacher model's output distribution~\cite{song2020lightpaff,liang2020mixkd,zhang2023not}, hidden states~\cite{jiao-etal-2020-tinybert,sun2019patient}, or attention scores~\cite{wang2020minilm,wang2021minilmv2}. For text generation tasks, knowledge distillation is predominantly applied through two distinct methodologies: word-level~\cite{sanh2019distilbert,song2020lightpaff} and sequence-level approaches~\cite{kim2016sequence,vicuna2023,alpaca,gu2023minillm}. At the word level, the process involves minimizing the forward Kullback-Leibler divergence (KLD) between the output distributions of the student and teacher models at each token step, effectively using the teacher's output probabilities as a supervisory signal to refine the student's predictions incrementally. Conversely, the sequence-level approach entails training the student model directly on complete texts generated by the teacher, thereby facilitating the acquisition of the teacher's stylistic and structural characteristics across entire sentences.

\section{Experiment Details}
\subsection{Dataset}\label{append:dataset}
\setcounter{figure}{4} 
\begin{figure*}[ht]
\centering
  \centering
  \subfloat[]{\centering{\includegraphics[width=0.4\textwidth]{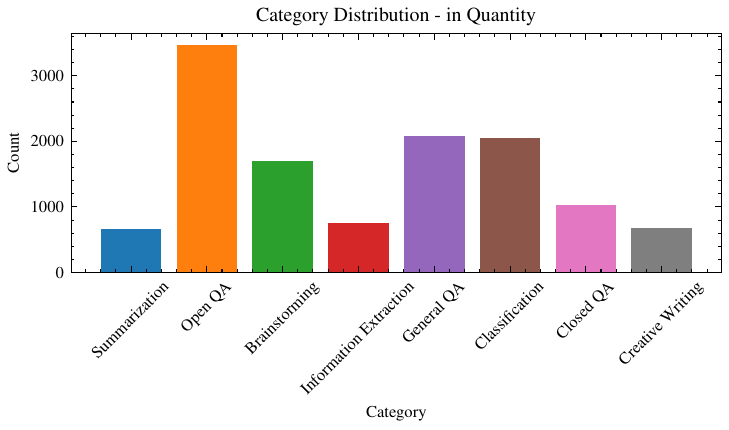} } } 
  \subfloat[]{\centering{\includegraphics[width=0.4\textwidth]{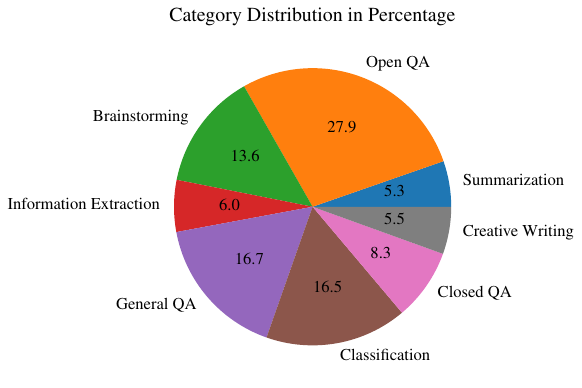} } }
  \caption{Bar and pie charts of the number (a) and corresponding percentage (b) of each category in the Dolly dataset.}
  \label{fig:dolly_dist}
\end{figure*}

\begin{figure*}[ht]
\centering
  \centering
  \includegraphics[width=0.9\textwidth]{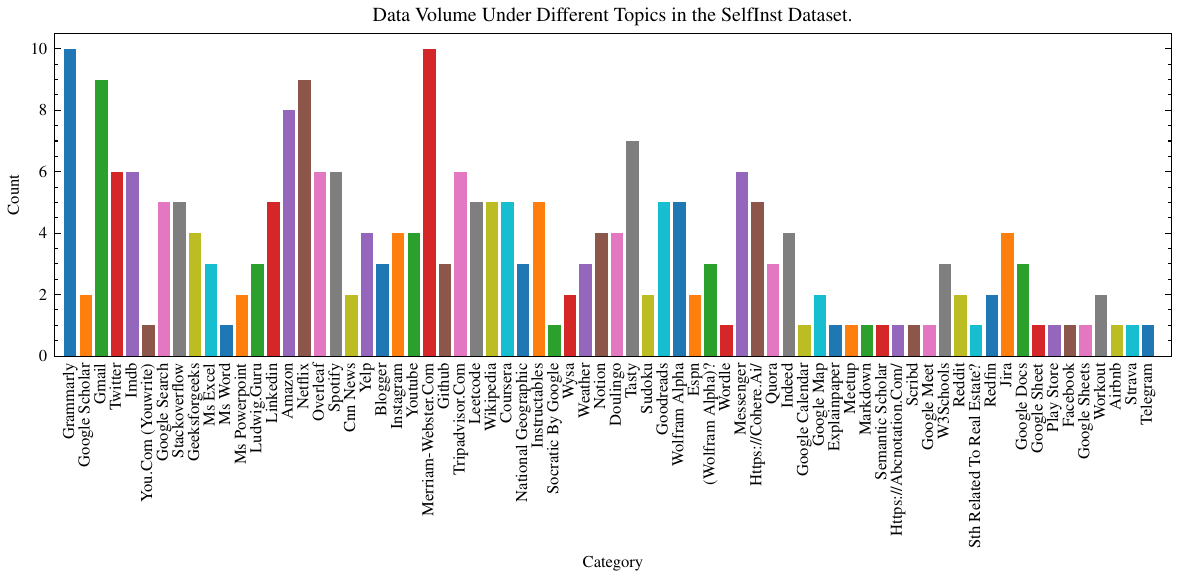} 
  \caption{Bar chart of the number of each category in the SelfInst dataset.}
  \label{fig:dolly_selfinst}
\end{figure*}
\begin{figure*}[ht]
\centering
  \centering
  \includegraphics[width=0.9\textwidth]{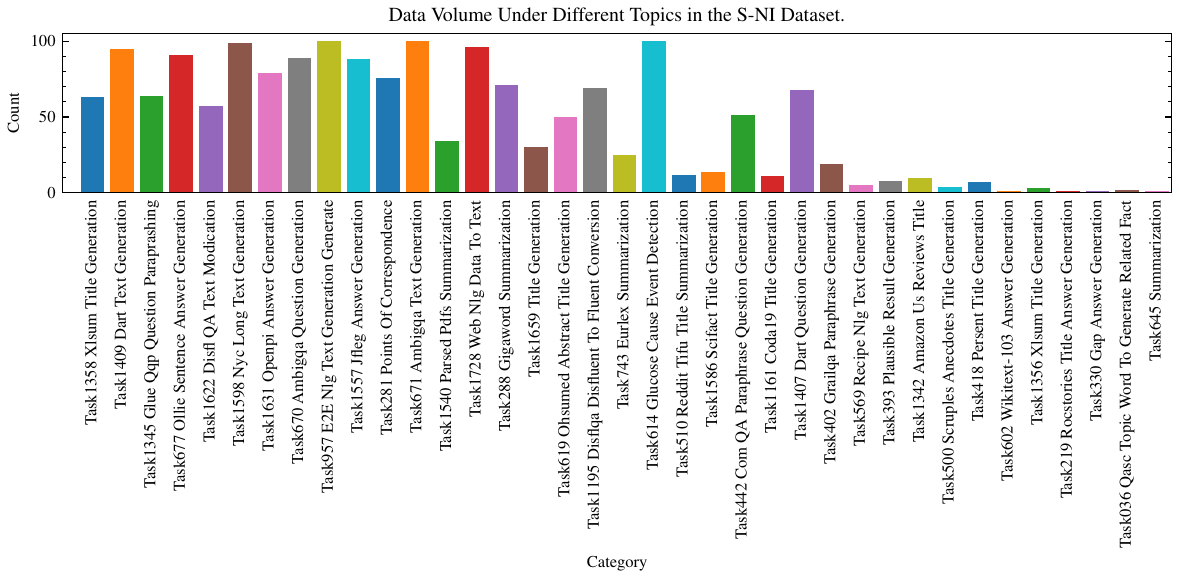}
  \caption{Bar chart of the number of each category in the S-NI dataset.}
  \label{fig:dolly_sinst}
\end{figure*}
Dolly (``databricks/databricks-dolly-15k'')\footnote{\url{https://huggingface.co/datasets/databricks/databricks-dolly-15k}} is an open-source collection of 15,000 high-quality human-generated prompt and response pairs designed for training and evaluating natural language processing models. This dataset contains over 15,000 records covering a range of instructional categories including brainstorming, classification, closed question answering (QA), generation, information extraction, open QA, and summarization. These categories were chosen to reflect different types of cognitive tasks that could be useful for training LLMs to respond in human-like manners across a variety of contexts. We plot the number of data samples and their corresponding percentage in Figure~\ref{fig:dolly_dist}.

SelfInst dataset\footnote{\url{https://github.com/yizhongw/self-instruct}} is designed to evaluate the practical utility of instruction-following models in user-oriented contexts. This dataset includes a diverse array of tasks accompanied by specific instructions, including tables, codes, or math equations. In total, it contains 252 distinct tasks, each associated with a unique instruction, aimed at testing the capability of models across a broad spectrum of applications. We show the number of data samples in test dataset from each category in Figure~\ref{fig:dolly_selfinst}.

S-NI dataset (Super-NaturalInstructions)~\cite{wang-etal-2022-super} is designed to test the generalization capabilities of LMs across a wide range of NLP tasks through declarative instructions. It includes over 1,600 unique tasks, encompassing diverse categories such as text classification, summarization, question answering, and more complex reasoning tasks. We draw the number of data samples used for evaluation in each category in Figure~\ref{fig:dolly_sinst}.

\subsection{Data Partition Strategy}\label{append:data_hete}
\setcounter{table}{2}
\begin{table*}
  \centering
  \begin{tabular}{clcccc}
\toprule
\multirow{2}{*}{Model} & \multirow{2}{*}{Method} & \multicolumn{3}{c}{Dataset} & \multirow{2}{*}{Model Size } \\ \cmidrule(lr){3-5}
                       &                         & Dolly & SelfInst & S-NI  &     \\ 
   \midrule
    \multirow{5}{*}{GPT-2}& FedAvg (1.5B) & 19.1\textsubscript{$\pm$.6} & 11.2\textsubscript{$\pm$.4} & 20.7\textsubscript{$\pm$.3} & 1.5B  \\
    \cmidrule(r){2-6}
    & Base (1.5B) & 7.2\textsubscript{$\pm$.1} & 5.5\textsubscript{$\pm$.3} & 5.8\textsubscript{$\pm$.1} & N/A\\
    & FedAvg (760M) & 18.0\textsubscript{$\pm$.5} & 10.1\textsubscript{$\pm$1.} & 17.1\textsubscript{$\pm$.3} & 760M  \\
      & FedAvg+PT (760M-1.5B) & 18.6\textsubscript{$\pm$.5} & 10.2\textsubscript{$\pm$.8} & 19.7\textsubscript{$\pm$.3} & 760M \\
    & FedPT (760M-1.5B) & \textbf{18.8}\textsubscript{$\pm$.4} & \textbf{11.2}\textsubscript{$\pm$.4} & \textbf{20.3}\textsubscript{$\pm$.2} & 760M  \\
    \bottomrule
  \end{tabular}
  \caption{Evaluation results on Dirichlet distribution. We report the average and standard deviation of Rouge-L scores across 5 random seeds. Higher values indicate better performance.}
  \label{tab:dirichlet_rouge}
\end{table*}
Federated fine-tuning LLMs involves tuning algorithms across multiple decentralized devices or servers holding local data samples, which are usually not identically distributed. This scenario frequently occurs in real-world applications, where data naturally varies across devices due to geographic diversity and user behavior. For example, diverse devices might engage in distinct activities like open-domain QA and creative writing. In this case, the format and content of instructions can be significantly different. For instance, QA tasks often focus on factual queries and responses, whereas creative writing tasks require guidelines for crafting engaging and imaginative narratives.

\begin{figure}[ht]
\centering
  \centering
  \includegraphics[width=0.75\linewidth]{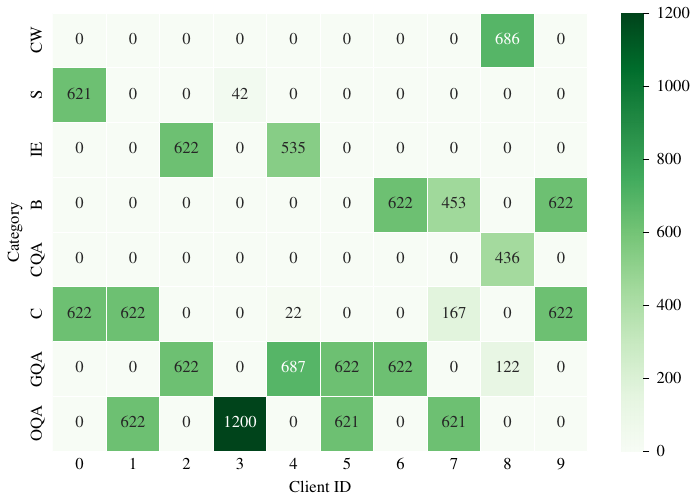}
  \caption{The pathological non-IID distribution of instruction categories distribution across devices. Categories: creative writing(CW), summarization(S), information extraction(IE), brainstorming(B), closed QA(CQA), classification(C), general QA(GQA), open QA(OQA).}
  \label{fig:data_path_dist}
\end{figure}

\begin{figure}[ht]
\centering
  \centering
  \includegraphics[width=0.75\linewidth]{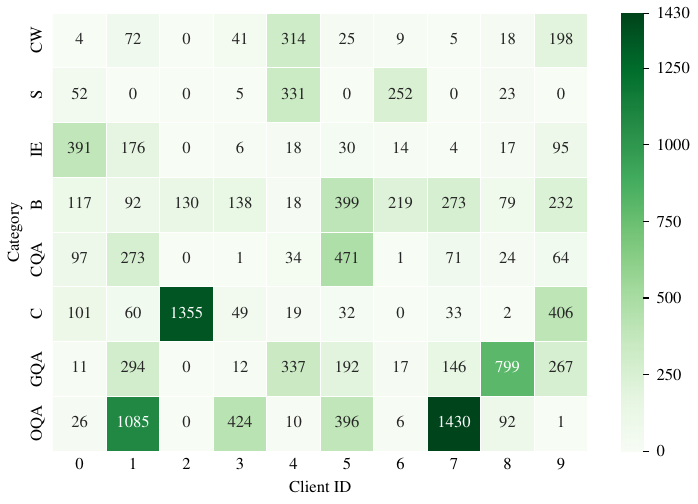}
  \caption{The Dirichlet non-IID distribution of instruction categories across devices.}
  \label{fig:data_diri_dist}
\end{figure}

To simulate an FL setup, we employ two data partition strategies, pathological non-IID~\cite{mcmahan2017communication} and Dirichlet non-IID~\cite{hsu2019measuring}. Specifically, we first sort the data from the Dolly dataset by categories. Then we randomly partition the dataset into 10 shards. For pathological non-IID distribution, each shard contains an equal number of samples and exclusively represents two specific categories. For Dirichlet distribution, the data from the same category are distributed among shards following Dirichlet distribution with concentration parameter 0.5. These segmentation strategies followed a commonly used partitioning method in~\cite{zhang2024towards,he2020fedml,lai2022fedscale,Shepherdgithub}, which led to a non-IID data distribution among the devices with imbalanced categories of instructions, mirroring a typical real-world FL data distribution. 
% We adopt the data partitioning approach by~\cite{zhang2024towards}, applied to the Databricks-Dolly-15K dataset, to effectively simulate non-IID conditions and assess the robustness of our models within federated environments. The instruction categories of Databricks-Dolly-15K include open QA, classification, brainstorming, summarization, general QA, closed QA, information extraction, and creative writing. We first divide the entire Databricks-dolly-15k dataset into ten shards using a widely adopted partitioning technique in t~\cite{he2020fedml,lai2022fedscale,zhang2023fed}, and then assign each shard to an individual device. 
Figures~\ref{fig:data_path_dist},~\ref{fig:data_diri_dist} depict the distribution of instruction categories within each device's dataset, with the former showing the pathological distribution and the latter displaying the Dirichlet distribution, respectively. As shown in Figure~\ref{fig:data_path_dist}, for pathological distribution, each device has imbalanced instruction categories with some categories completely missing. For Dirichlet distribution, Figure~\ref{fig:data_diri_dist} illustrates that each device has an imbalanced distribution of instruction categories and a varying total number of samples. These imbalances mirror real-world conditions, where individual users often encounter a skewed variety of instructions, reflecting their unique usage patterns and preferences. 

We apply the Dirichlet distribution to the training dataset and present the evaluation results for the GPT-2 models in Table~\ref{tab:dirichlet_rouge}. The results indicate that FedPT outperforms the federated fine-tuning small model, such as FedAvg (760M) and FedAvg+PT (760M-1.5B), and performs comparably to federated fine-tuning large models like FedAvg (1.5B).  
%%%%%%%%%%%%%%%%%%%%%%%
%%%%%%%%%%%%%%%%%%%%%%%%
%%%%%%%%%%%%%%%%%%%%%%%
%%%%%%%%%%%%%%%%%%%%%%%%

%------------------------------------
\subsection{ GPT-4 Evaluation Configuration}\label{append:gpt4score}
%------------------------------------
We use the Rouge-L scores and GPT-4 feedback scores to evaluate the model-generated responses. These approaches ensure a more balanced and comprehensive evaluation of the model's ability to produce high-quality, contextually appropriate text. Following the same evaluation approach in~\cite{gu2023minillm,shen-etal-2023-large}, we utilize GPT-4 as a judge to compare model-generated responses with the ground truth answers, assigning scores from 1 to 10 for both sets of responses. We call the GPT-4 Turbo API\footnote{API version of 2024-04-09.} with the temperature $=0.7$. The evaluation prompt used for GPT-4 is illustrated in Figure~\ref{fig:auto_eval_prompt}. We calculate the ratio of the total scores of model-generated responses and the ground truth answers. We select the seed closest to the average Rouge-L score and then report its GPT-4 feedback score. For Dolly and SelfInst datasets, we evaluate all the responses. For the S-NI dataset, we randomly select 200 responses for evaluation. The results are summarized in Table~\ref{tab:gpt4_eval_llama} and Table~\ref{tab:gpt4_eval_gpt2}. These tables demonstrate that FedPT can achieve performance comparable to the direct tuning of the large models in the FL setting. 
\begin{figure}[h!]
\centering
\begin{tcolorbox}[colback=gray!20,  % Light gray background
                  colframe=black,  % Black border
                  boxrule=0.5mm,   % Border thickness
                  width=0.45\textwidth, % Box width
                  rounded corners] % This option rounds the corners

You are a helpful and precise assistant for checking the quality of the answer.\\

% \vspace{0.8em}
[Instruction]\\
% \textbf{\#\#\# Instruction:}\\
\{instruction\}\\
{[Input]}\\
% \textbf{\#\#\# Input:}\\
\{input\}\\
{[}The Start of Assistant 1’s response{]}\\
\{answer 1\}\\
{[}The End of Assistant 1’s Answer{]}\\
{[}The Start of Assistant 2’s response{]}\\
\{answer 2\}\\
{[}The End of Assistant 2’s Answer{]}\\
{[System]} \\
We would like to request your feedback on the performance of two AI assistants in response to the user instruction and input displayed above.\\
Please rate the helpfulness, relevance, accuracy, and level of detail of their responses. Each assistant receives an overall score on a scale of 1 to 10, where a higher score indicates better overall performance.\\
Please first provide a comprehensive explanation of your evaluation, avoiding any potential bias and ensuring that the order in which the responses were presented does not affect your judgment. \\
Then, output two lines indicating the scores for Assistant 1 and 2, respectively. \\
Output with the following format: \\
Evaluation evidence: $<$your evaluation explanation here$>$ \\
Score of the Assistant 1: $<$score$>$ \\
Score of the Assistant 2: $<$score$>$

\end{tcolorbox}
\caption{GPT-4 evaluation prompt.}\label{fig:auto_eval_prompt}
\end{figure}
% 
% \begin{table}[htbp]
%     \centering
%     \begin{minipage}{0.6\textwidth}
%         \centering
%         \caption{Evaulation results of LLaMA by GPT-4 feedback. Higher scores indicate better performance.}
%         \begin{tabular}{lccc}
%             \toprule
%             \textbf{Method} & \textbf{Dolly} & \textbf{SelfInst} & \textbf{S-NI} \\
%             \midrule
%             FedAvg (13B) & 65.4 & 59.5 & 61.8 \\
%             FedAvg (7B) & 57.7 & 52.1 & 50.7 \\
%             FedAvg+PT (7B-13B)  & 63.6 & 56.4 & 59.0 \\
%             FedPT (7B-13B)& 65.4 & 60.3 & 61.6 \\

%             \bottomrule
%         \end{tabular}
%     \end{minipage}%
%     % \hspace{0.04\textwidth}
%     % \begin{minipage}{0.45\textwidth}
%     %     \centering
%     %     \caption{GPT-2 Evaluation Results}
%     %     \begin{tabular}{lccc}
%     %         \toprule
%     %         \textbf{Method} & \textbf{Dolly} & \textbf{SelfInst} & \textbf{S-NI} \\
%     %         \midrule
%     %         FedAvg (1.5B) & 35.7 & 29.1 & 29.2 \\
%     %         FedAvg (760M) & 30.3 & 26.1 & 24.5 \\
%     %         FedAvg+PT  & 34.4 & 28.2 & 26.8 \\
%     %         FedPT  & 34.8 & 28.8 & 27.8 \\
%     %         \bottomrule
%     %     \end{tabular}
%     % \end{minipage}
%     % \caption{Evaluation results by GPT-4 feedback. Higher values indicate better performance.}
% \end{table}
\begin{table}[]
    \centering
    \begin{tabular}{lcccc}
    \toprule
        Method & Dolly & SelfInst & S-NI \\
        \midrule
          FedAvg (1.5B) & 35.7 & 29.1 & 29.2   \\
   \midrule
    % & Base (1.5B) &  &  &  & N/A\\
     FedAvg (760M) & 30.3 & 26.1 & 24.5   \\
       FedAvg+PT (760M-1.5B)& 34.4 & 28.2 & 26.8  \\
     FedPT  (760M-1.5B)& \textbf{34.8} & \textbf{28.8} & \textbf{27.8}  \\
     \bottomrule
    \end{tabular}
\caption{Evaluation results by GPT-4 feedback on GPT-2. Higher scores indicate better performance.}\label{tab:gpt4_eval_gpt2}
\end{table}

% \begin{table}
%  \caption{Evaluation results by GPT-4 feedback. Higher values indicate better performance.}\label{tab:gpt_eval}
%   \centering
%   \begin{tabular}{@{}clccc@{}}
% \toprule
% \multirow{2}{*}{Model} & \multirow{2}{*}{Method} & \multicolumn{3}{c}{Dataset}  \\ \cmidrule(lr){3-5}
%                        &                         & Dolly & SelfInst & S-NI  \\
%     \midrule
%     \multirow{4}{*}{LLaMA} & FedAvg (13B) & {65.4} & {59.5} & 61.8 \\
%     \cmidrule(r){2-5}
%     % &Base (13B) &  &  &  & N/A \\
%     &FedAvg (7B) & 57.7 & 52.1 & 50.7 \\   
%     & FedAvg+PT (7B-13B) & 63.6 & 56.4 & 59.0\\
%     & FedPT (7B-13B)& \textbf{65.4} & \textbf{60.3} & \textbf{61.6} \\
    
%    \midrule
%     \multirow{4}{*}{GPT-2}& FedAvg (1.5B) & 35.7 & 29.1 & 29.2   \\
%     \cmidrule(r){2-5}
%     % & Base (1.5B) &  &  &  & N/A\\
%     & FedAvg (760M) & 30.3 & 26.1 & 24.5   \\
%       & FedAvg+PT (760M-1.5B)& 34.4 & 28.2 & 26.8  \\
%     & FedPT  (760M-1.5B)& \textbf{34.8} & \textbf{28.8} & \textbf{27.8}  \\
%     \bottomrule
%   \end{tabular}
% \end{table}

%%%%%%%%%%%%%%%%%%%%%%%%%%%%%%%%%%%%%
\subsection{Automatic Evaluation Details}
%%%%%%%%%%%%%%%%%%%%%%%%%%%%%%%%%%%%%
In our evaluation process, we extract responses from each model by setting the temperature to 1, limiting responses to a maximum length of 512, and employing random seeds \{10, 20, 30, 40, 50\}. Following the previous works~\cite{taori2023stanford,gu2023minillm}, we utilize a prompt wrapper illustrated in Figure~\ref{fig_prompt} to reformat each pair of instruction-response into a sentence.
\begin{figure}[t]
\centering
\begin{tcolorbox}[colback=gray!20,  % Light gray background
                  colframe=black,  % Black border
                  boxrule=0.5mm,   % Border thickness
                  width=0.45\textwidth, % Box width
                  rounded corners] % This option rounds the corners
Below is an instruction that describes a task.\\
Write a response that appropriately completes the request.

{[Instruction]}\\
% \textbf{\#\#\# Instruction:}\\
\{instruction\}\\
% \vspace{0.8em}

{[Input]}\\
\{input\}\\
% \vspace{0.8em}

[Response]
% \textbf{\#\#\# Response:}
\end{tcolorbox}
\caption{The prompt wrapper for training and evaluation.}\label{fig_prompt}
\label{fig:promptwrapper}
\end{figure}

\begin{table}[t]
  \centering
  \begin{tabular}{l|c|c}
    \toprule
    Hyperparameter & GPT-2 & LLaMA  \\
    \midrule
    Precision & Float16 & Float16\\
    Number of local epochs &2 &2   \\
    Total round & 20 & 20\\
    Training Batch size &64 &64  \\
    Learning rate &1 &1 \\
    Weight decay&0.01 & 0.01\\
    Max sequence length & 512&512  \\
    Knowledge distillation data size &128 &512  \\
    Knowledge distillation batch size &16 &32  \\
    Knowledge distillation iterations &8 &16 \\
    \bottomrule
  \end{tabular}
  \caption{Hyper-parameters for proxy-tuning task-specific models.} \label{table:lora_parameters}
\end{table}

\begin{table*}[t]
  
  \centering
  \begin{tabular}{crc|c|c|c}
    \toprule
    Model &\#Size & Rank  & \makecell{Trainable Param} & \makecell{LoRA Size}  & \makecell{Trainable Fraction} \\
    \midrule
    \multirow{2}{*}{GPT-2} & \multirow{1}{*}{760M} & 4 &  0.7 M& 1.4 MB & $0.09 \%$ \\
                           & \multirow{1}{*}{1.5B} & 4 &  1.2 M & 2.4 MB & $ 0.08\%$ \\
    \cmidrule(r){2-6}
    \multirow{2}{*}{LLaMA} & \multirow{1}{*}{7B} & 8 &  4.2 M & 8.1 MB & $0.06\% $\\
                           & \multirow{1}{*}{13B} & 8 &  6.5 M & 12.5 MB & $0.05\% $ \\    
                           & \multirow{1}{*}{30B} & 8 &  12.78 M & 25.6 MB & $0.04\% $ \\    
    \bottomrule
  \end{tabular}
  \caption{LoRA training configuration on user devices.}\label{tab:resource}
\end{table*}

\subsection{Hyper-parameters} \label{append:hyper-para}
The specific configurations are documented in Table~\ref{table:lora_parameters}.
During evaluation, we consistently generate responses using greedy search with unrestricted sampling. The Top-p ratio is set to $1.0$ and the temperature to $1.0$. The maximum generation length is capped at 512 tokens. Evaluation batch sizes are 32 for the GPT-2 model and 8 for the LLaMA model, respectively.

%%%%%%%%%%%%%%%%%
\subsection{LoRA Configuration}\label{append:lora_parameters}
%%%%%%%%%%%%%%%%%
We apply LoRA to the attention layer for GPT-2 model and ``q\_proj'', ``v\_proj'' layers for LLaMA model to enhance adaptation capabilities, using the Adam optimizer for effective training. We set the rank of LoRA to be 4 and 8 for GPT-2 and LLaMA, respectively. This only yields 4.2 M trainable parameters with size 8.1 MB for LLaMA-7B model, which is affordable for many user devices. The overall LoRA training configuration for different models can be found in Table~\ref{tab:resource}.

\subsection{Hardware and Library}
We conduct the experiment on the Ubuntu (22.04.4 LTS) server equipped with 4 A6000 GPUs. Each GPU has 48 GB VRAM. The training scripts were implemented using Pytorch 2.0.1~\cite{paszke2019pytorch}. To accelerate the experiment's progress, we also employ popular open-sourced third-party packages, including transformer 4.36.0.dev0~\cite{wolf-etal-2020-transformers}, deepspeed 0.14.0~\cite{rasley2020deepspeed}, accelerate 0.29.2~\cite{accelerate}, nltk 3.8.1~\cite{bird2009natural}, sentencepiece 0.2.0~\cite{kudo2018sentencepiece}, and datasets 2.81.0~\cite{lhoest-etal-2021-datasets}. For LoRA local training, we implement the low-rank model update using PEFT package~\cite{peft}. For all experiments, we adopt the Python 3.10 interpreter and CUDA version 11.4.

%------------------------------
\section{Result Analysis}
%-------------------------------
\subsection{Evaluation of Tokens Most Influenced by Proxy-Tuning}\label{append:top_tokens}

We aim to investigate which tokens are the most influenced by FedPT. To this end, we calculate the frequency of each token in the generated responses for  GPT-2-1.5B to its FedPT version. Table~\ref{tab:tokens} summarizes the 8 tokens whose occur frequency is most increased from GPT-2-1.5B to its FedPT. We can see that these tokens are more contributing to reasoning and style. These findings are consistent with the hypothesis that instruction-tuning mainly influences reasoning and style, rather than increasing the model’s knowledge \cite{gudibande2023false}.
\begin{table*}[t]
    \centering
    \begin{tabular}{cc|cc|cc}
        \toprule
        \multicolumn{2}{c}{Dolly} & \multicolumn{2}{c}{SelfInst} & \multicolumn{2}{c}{S-NI}\\
        \midrule
        {Token} & {Top Context}& {Token} & {Top Context}& {Token} & {Top Context} \\
        \midrule
        is & is one of the & equal & is equal to & because & because he is\\
        a & it can be a & well & as well as & he & when he was\\
        can & can be used to & was & said it was & in & facts specified in\\
        popular & the most popular & do & need to do & they & they are not\\
        most & of the most & it & and it will & changed & entity changed from\\
        known & is known for & into & into something new & has & an individual has\\
        when & when I was & nothing & there is nothing & were & and I were \\
        many & there are many & when & when a change & is & there is a\\
        \bottomrule
    \end{tabular}
    \caption{For the three datasets, the 8 tokens whose occur frequency increased the most from GPT-2-1.5B to its FedPT version. {Top Context} shows the most common 3-gram that the word occurs in.}\label{tab:tokens}
\end{table*}
\begin{table*}[ht]
  \centering
  \begin{tabular}{clcccccc}
    \toprule
    \multirow{2}{*}{Model} & \multirow{2}{*}{Method} & \multicolumn{2}{c}{Dolly} & \multicolumn{2}{c}{SelfInst}  & \multicolumn{2}{c}{S-NI}\\
    \cmidrule(r){3-8}
    & &  Dist-3 & Dist-4& Dist-3 & Dist-4 & Dist-3 & Dist-4\\
    \midrule
    \multirow{5}{*}{LLaMA} & FedAvg (13B) & $ 96.4$ & $ 99.3$ & $ 97.5$ & $99.4 $ & $ 93.2$ & $98.0 $ \\
    \cmidrule(r){2-8}
     & Base (13B) & $ 95.0$ & $ 99.1$ & $96.3 $ & $ 99.3$ & $89.9 $ & $97.9 $ \\
    & FedAvg (7B) & $96.6 $ & $ 99.3$ & $97.7 $  & $ 99.5$ & $ 93.5$& $98.4 $\\
    & FedAvg+PT & $ 97.1$ & $99.4$ & $98.0 $ & $ 99.6$ & $ 94.3$ & $ 98.5$  \\   
     & FedPT & $97.2 $ & $99.4 $ & $ 98.0$ & $99.5 $ & $ 93.8$ & $98.2 $ \\
     
   \midrule

    \multirow{5}{*}{GPT-2} & FedAvg (1.5B) & $97.0 $ & $99.4 $ & $ 98.1$ & $ 99.5$ & $94.7 $ & $98.5 $ \\
    \cmidrule(r){2-8}
    & Base (1.5B) & $ 96.3$ & $99.4$ & $ 97.0$ & $ 99.4$ & $92.2$ & $95.6 $  \\
    & FedAvg (760MB) & $ 97.0$ & $99.4 $ & $98.2 $  & $ 99.6$ & $94.8$ & $98.6$ \\
    & FedAvg+PT (760M-1.5B)  & $97.0 $ & $ 99.4$ & $ 98.3$  & $ 99.6$ & $94.7 $ & $ 98.3$ \\
    & FedPT (760M-1.5B) & $ 97.4$ & $ 99.5$ & $ 98.5$ & $ 99.6$ & $ 93.3$ & $97.5 $ \\
    \bottomrule
  \end{tabular}
  \caption{The distinct 3-grams and 4-grams (Dist-3 and Dist-4) on the test sets. FedPT preserves generation diversity.}
  \label{tab:diversity}
\end{table*}
\subsection{Generation Diversity}
\label{append:generation diversity}

Table~\ref{tab:tokens} shows that the occurrence frequency of certain tokens increased from LLaMA-7B to its proxy-tuned version, potentially affecting generation diversity. To investigate this impact, we conducted experiments on distinct n-grams (Dist-3 and Dist-4) diversity, a widely used metric to measure the generation diversity of an LM~\cite{li2016diversity} (see Appendix~\ref{append:diversity} for more details). As shown in Table~\ref{tab:diversity}, our algorithm maintains a high level of diversity despite the observed changes in token frequency.

%%%%%%%%%%%%%%%%%%%%%%%%%%%%%%%%%%%%
\subsection{Further Evaluation of $\bm{\alpha}$}\label{append:alpha_appendix}
\begin{figure*}[t]
\centering
    \subfloat[Dolly]{\centering{\includegraphics[width=0.3\textwidth]{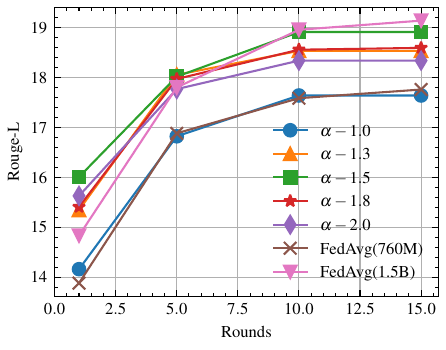} } }
    \subfloat[SelfInst]{\centering{\includegraphics[width=0.3\textwidth]{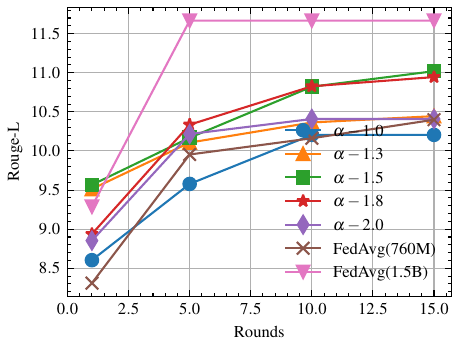} } }
    \subfloat[S-NI]{\centering{\includegraphics[width=0.3\textwidth]{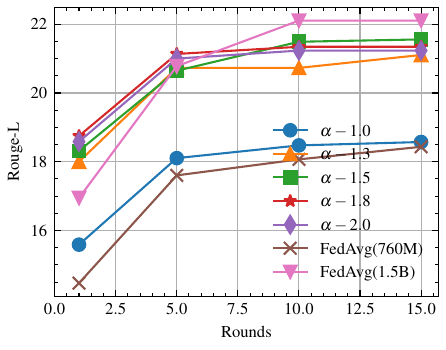} } }
    % \subfloat[Dolly]{\centering{\includegraphics[width=0.3\textwidth]{./Figures/fedpt/llama-alpha/dolly_Rouge-L.pdf} } }
    % \subfloat[SelfInst]{\centering{\includegraphics[width=0.3\textwidth]{./Figures/fedpt/llama-alpha/self_inst_Rouge-L.pdf} } }
    % \subfloat[S-NI]{\centering{\includegraphics[width=0.3\textwidth]{./Figures/fedpt/llama-alpha/sinst_Rouge-L.pdf} } }
    \caption{Performance comparison of different $\alpha$ for FedPT on GPT-2 across different rounds for Dolly, SelfInst, and S-NI tasks. Higher Rouge-L scores indicate better performance.}
    \label{fig:effect_alpha_gpt2}%
\end{figure*}

%%%%%%%%%%%%%%%%%%%%%%%%%%%%%%%%%%%%
We use different $\alpha$ values in FedPT to investigate the effect of proxy-tuning weight $\alpha$. Specifically, we set $\alpha \in \{1.0, 1.3, 1.5, 1.8, 2.0\}$ to evaluate the GPT-2 model on three testing datasets at global rounds $\{1, 5, 10, 15\}$. For the LLaMA model, we evaluate it at global rounds $\{1, 5, 10, 15, 20\}$, using $\alpha$ values in the range $\{1.0, 1.5, 2.0\}$. The Rouge-L scores for LLaMA and GPT-2 are illustrated in Figure~\ref{fig:effect_alpha_llama} and Figure~\ref{fig:effect_alpha_gpt2}, respectively. From these figures, we can find that the value $\alpha$ plays a crucial role in determining the behavior of the model. As $\alpha$ increases, the influence of the fine-tuned small model on the predictions becomes more pronounced, leading to more substantial deviations from the pre-trained large model's behavior. Conversely, as $\alpha$ decreases, the predictions tend to align more closely with the original pre-trained large model, resulting in a more stable and conservative output. Therefore, in practice, we need to carefully choose an appropriate $\alpha$ for the specific downstream task.

% This demonstrates the importance of carefully tuning $\alpha$ to balance the trade-off between leveraging fine-tuning adjustments and maintaining the stability of the pre-trained model's predictions.  As shown in Equation~\ref{eq: proxy-tuning}, the generated logit follows $g_{\bm{\theta}_{{l}}}+\alpha ( g_{\Bar{\bm{\theta}}_{{s}}} - g_{{\bm{\theta}}_{{s}}})$. Intuitively, larger $\alpha$ magnifies the influence of the difference between the fine-tuned small model and the pre-trained small model, making the predictions more responsive to the fine-tuning adjustments. Conversely, a smaller $\alpha$ results in predictions that are more similar to the large pre-trained model, causing the predictions to closely adhere to the original large pre-trained model's behavior. 

% Figure~\ref{fig:effect_alpha_llama} shows the results of $\alpha\in\{1.0, 1.5, 2.0\}$ using LLaMA on the three datasets. Additionally, we provide the results using GPT-2 in Appendix~\ref{append:alpha}.
% 
%---------------------------------------
%---------------------------------------
\subsection{Further Analysis in Scaling Law}
\begin{figure*}[ht]
\centering
    \subfloat[Dolly]{\centering{\includegraphics[width=0.3\textwidth]{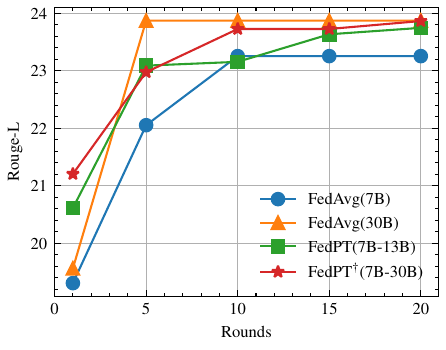} } }
    \subfloat[SelfInst]{\centering{\includegraphics[width=0.3\textwidth]{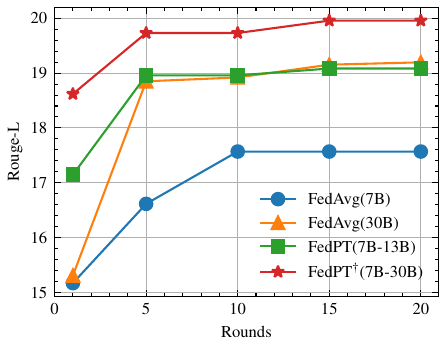} } } 
    \subfloat[S-NI]{\centering{\includegraphics[width=0.3\textwidth]{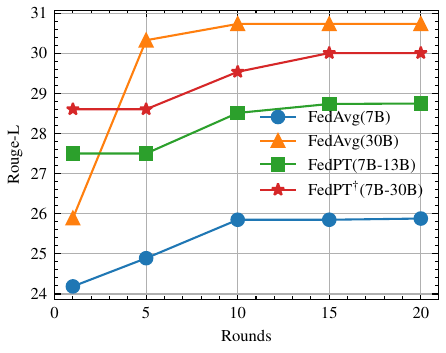} } }\\
    \subfloat[Dolly]{\centering{\includegraphics[width=0.3\textwidth]{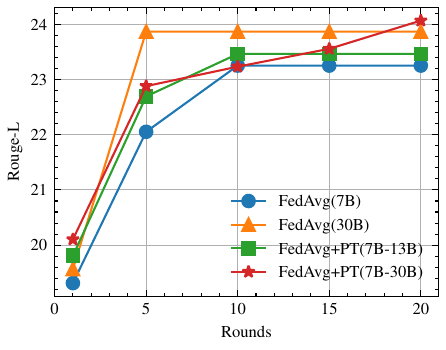} } }
    \subfloat[SelfInst]{\centering{\includegraphics[width=0.3\textwidth]{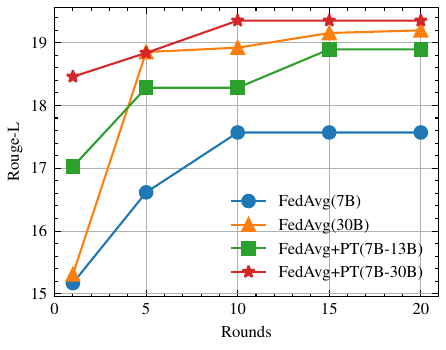} } }
    \subfloat[S-NI]{\centering{\includegraphics[width=0.3\textwidth]{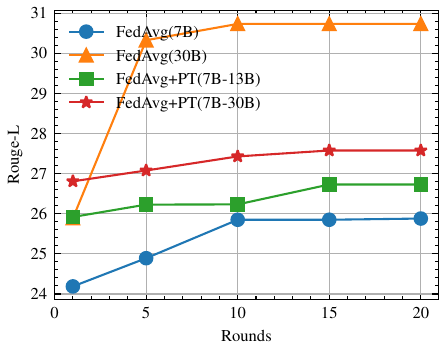} } }
    \caption{The scaling law of proxy-tuned models in the LLaMA family models. (a, b, c) are the results of FedPT. (d, e, f) are the results of FedAvg+PT. FedPT{$^\dagger$} (7B-30B) uses the fine-tuned 7B model from FedPT (7B-13B) to proxy-tune the LLaMA 30B. }
    \label{fig:scalling_llama}%
\end{figure*}
In this section, we analyze the performance of FedPT and FedAvg+PT at various scales for LLaMA models and show the results in Figure~\ref{fig:scalling_llama}. FedPT$^\dagger$(7B-30B) uses the fine-tuned 7B model from FedPT (7B-13B) to proxy-tune the LLaMA 30 model. As illustrated in Figure~\ref{fig:scalling_llama}, we observe performance enhancements for both FedPT and FedAvg+PT as the model size increases. The trend lines for both methods display a clear positive correlation between model size and performance, validating the scaling law: larger models tend to yield better results. 

% Notably, FedPT consistently outperforms FedAvg+PT, reinforcing the advantage of our proposed method as the number of model parameters increases. 

\subsection{ Details about Generation Diversity Metrics}\label{append:diversity}

Dist-n is calculated as a fraction $N/C$, where $N$ represents the number of distinct n-grams in the generated responses and $C$ denotes the total number of generated n-grams. We report the average values across 5 seeds in Table~\ref{tab:diversity}.

%%%%%%%%%%%%%%%%%%%%%%%%%%%%%%%%%%%%
\section{Supporting Plots}\label{append:category}
%%%%%%%%%%%%%%%%%%%%%%%%%%%%%%%%%%%%
In this section, we analyze the Rouge-L score and BLEU score among different category distributions for different models tuned by FedPT. Here, we only show the results of GPT-2(760M-1.5B) and LLaMA (7B-13B).

\subsection{Plots for Category Scores of Dolly} 
\begin{figure}[ht]
    \centering\subfloat[Global Round 1]{{\includegraphics[width=0.46\textwidth]{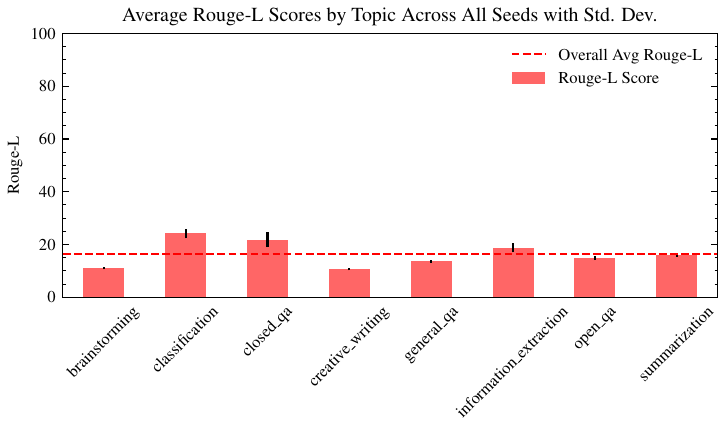} } }\\
    \centering\subfloat[Global Round 1]{{\includegraphics[width=0.46\textwidth]{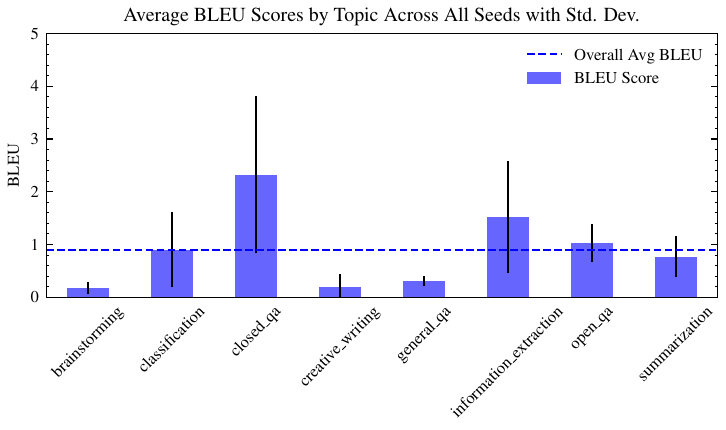} } }\\
    \centering\subfloat[Global Round 15]{{\includegraphics[width=0.46\textwidth]{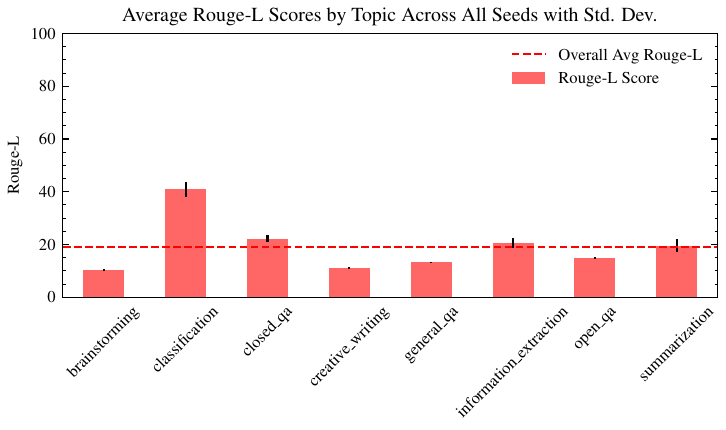} } }\\
    \centering\subfloat[Global Round 15]{{\includegraphics[width=0.46\textwidth]{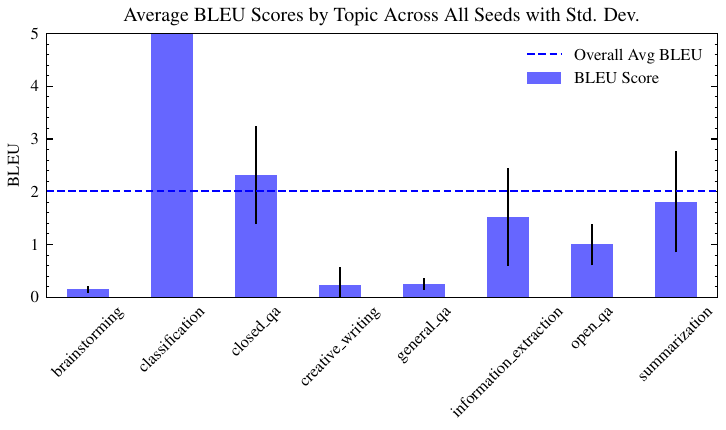} } }
    \caption{Rouge-L score distribution across different categories of Dolly dataset at global communication round 1 (a) and 15 (b) for FedPT. Model:GPT-2.}
    \label{fig:category_dolly}
\end{figure}

\begin{figure}[ht]
    \centering\subfloat[Global Round 1]{{\includegraphics[width=0.46\textwidth]{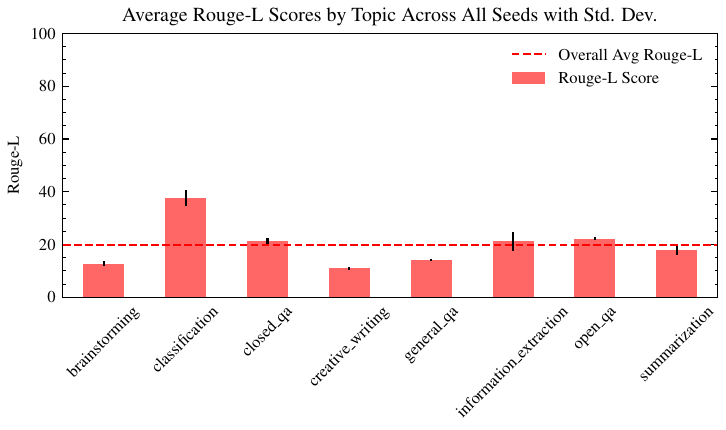} } }
    \\
    \centering\subfloat[Global Round 1]{{\includegraphics[width=0.46\textwidth]{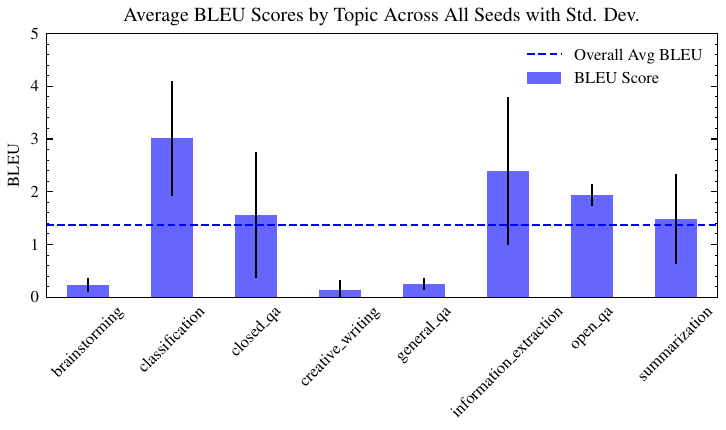} } }
    \\
    \centering\subfloat[Global Round 20]{{\includegraphics[width=0.46\textwidth]{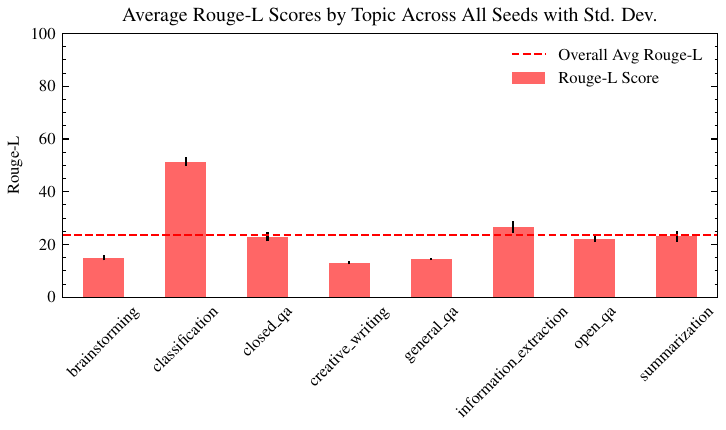} } }
    \\
    \centering\subfloat[Global Round 20]{{\includegraphics[width=0.46\textwidth]{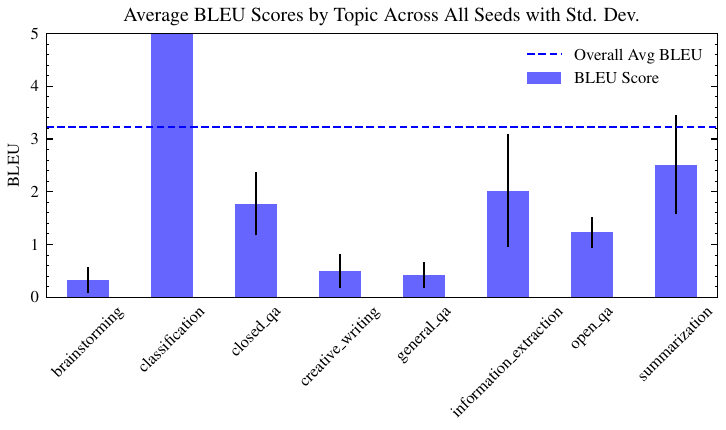} } }
    \\
    \caption{Rouge-L score distribution across different categories of Dolly dataset at global communication round 1 (a) and 20 (b) for FedPT. Model: LLaMA.}
    \label{fig:category_dolly_llama}%
\end{figure}

In Figure~\ref{fig:category_dolly}, \ref{fig:category_dolly_llama}, we plot the Rouge-L score and BLEU score from different categories of the Dolly dataset at round 1, 15 on GPT-2 model and round 1, 20 on LLaMA model. From the figures, we can find the scores for most categories are continually improving with global rounds increasing. The category ``classification'' leads the most contribution during training. In global round 1, the performance across tasks is fairly uniform, hovering around an overall average Rouge-L score (indicated by the dashed line), with ``classification'' scoring notably higher. By global rounds 15 and 20, there is a clear shift in performance; ``classification'' peaks significantly above other tasks, suggesting an improvement in the system's capability to handle classification tasks, while the other tasks show varied but generally less substantial improvement. The error bars for Rounds 15 and 20 tend to be smaller across various tasks, suggesting a potential decrease in variability and enhanced consistency in the model's performance among different random seeds.

\subsection{Plots for Category Scores of SelfInst} 
\begin{figure*}[ht]
    \centering\subfloat[Global Round 1]{{\includegraphics[width=0.7\textwidth]{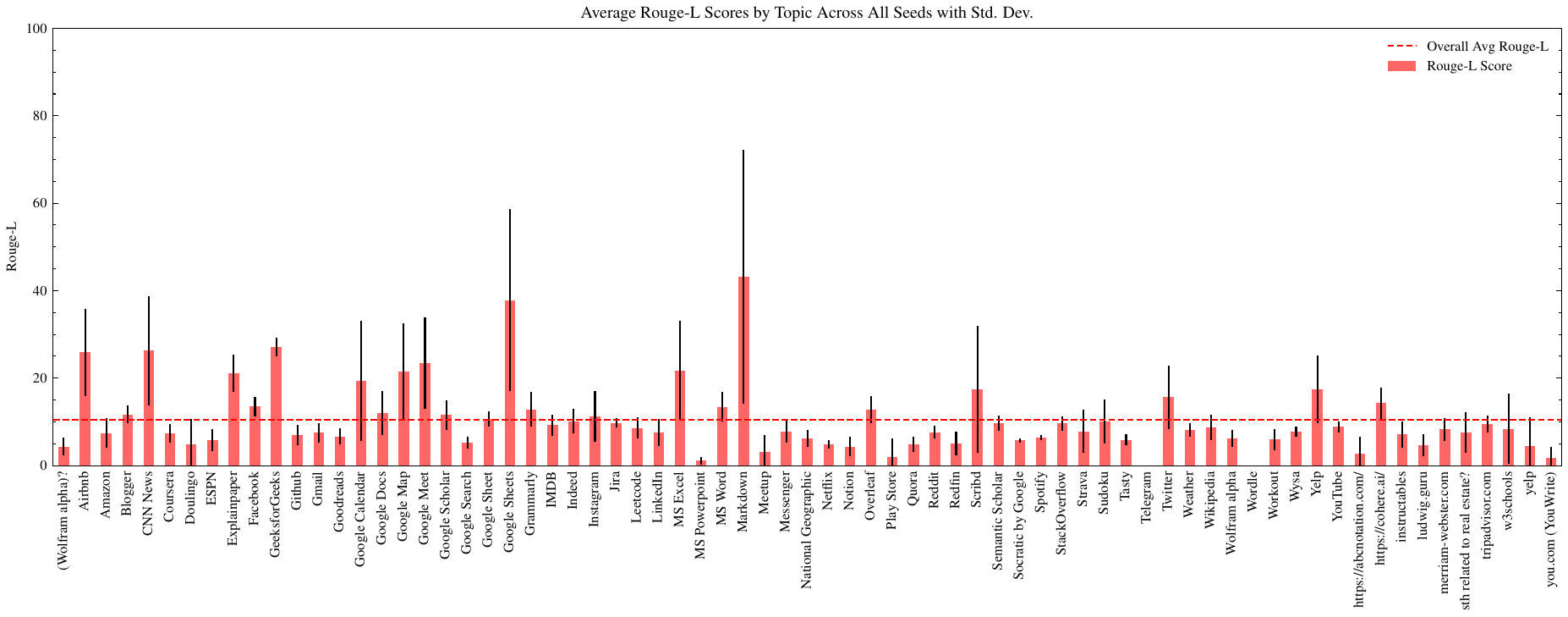} } }\\
    \centering\subfloat[Global Round 1]{{\includegraphics[width=0.7\textwidth]{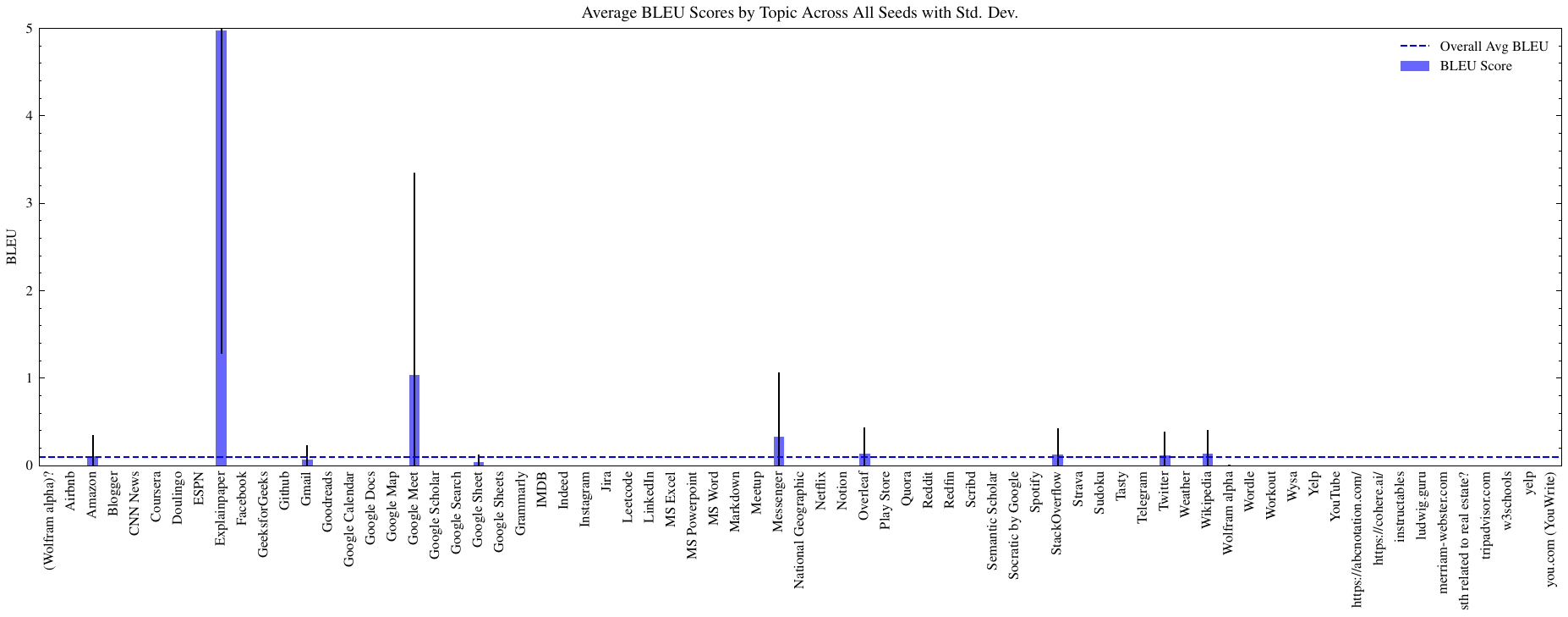} } }\\
    \centering\subfloat[Global Round 15]{{\includegraphics[width=0.7\textwidth]{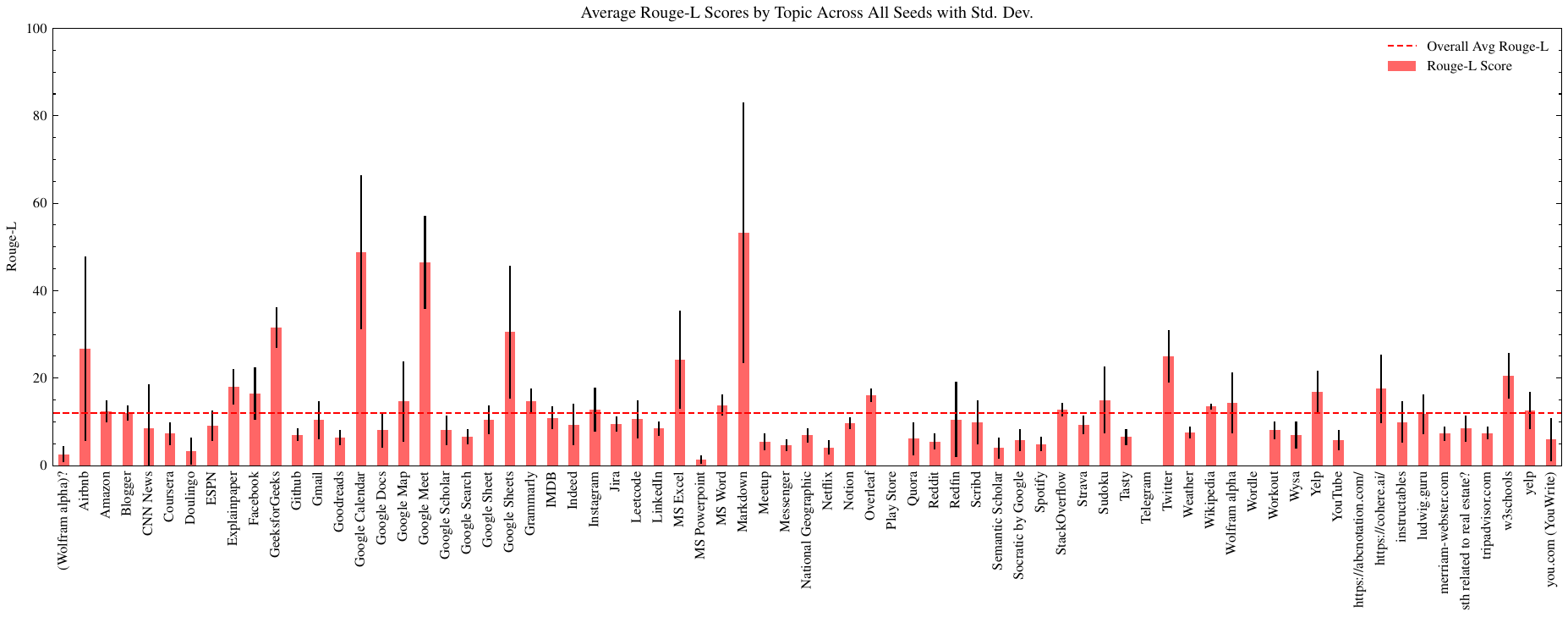} } }\\
    \centering\subfloat[Global Round 15]{{\includegraphics[width=0.7\textwidth]{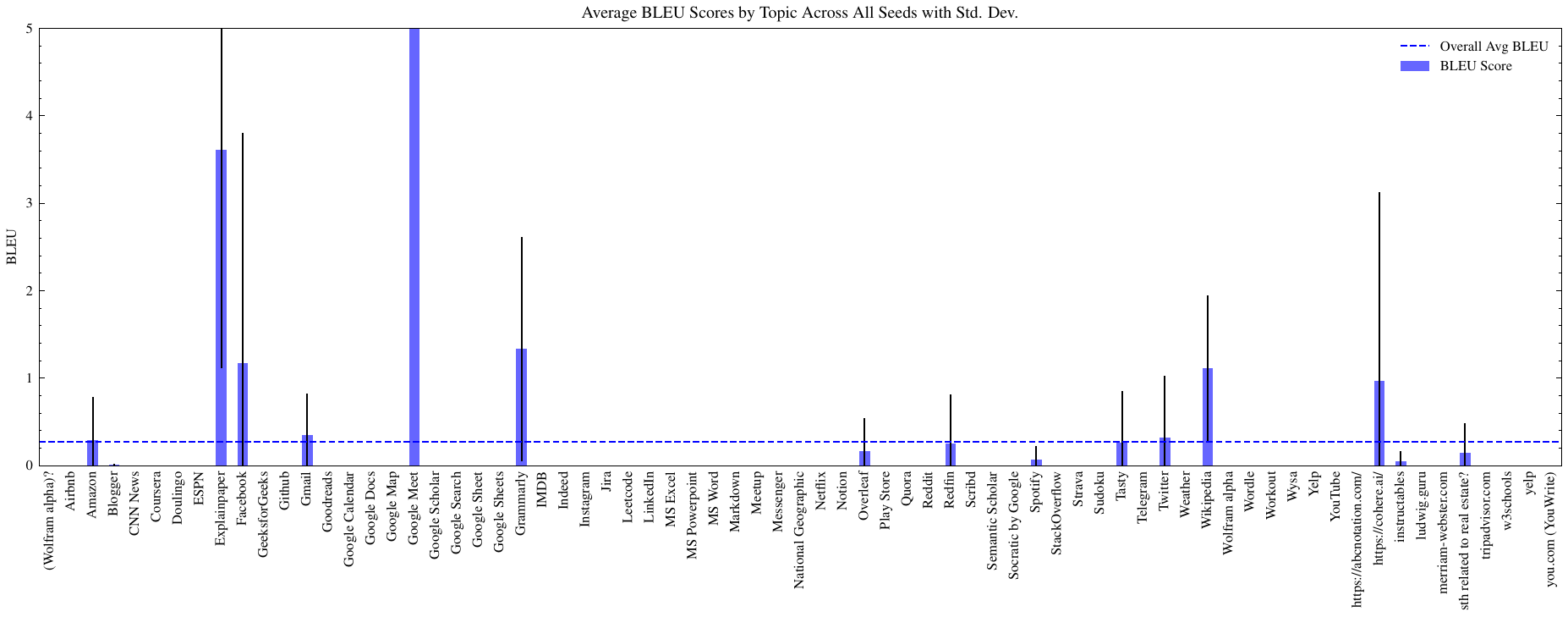} } }
    \caption{Rouge-L score distribution across different categories of SelfInst dataset at global communication round 1 (a) and 15 (b) for FedPT. Model:GPT-2.}
    \label{fig:category_self_inst}
\end{figure*}

\begin{figure*}[ht]
    \centering\subfloat[Global Round 1]{{\includegraphics[width=0.7\textwidth]{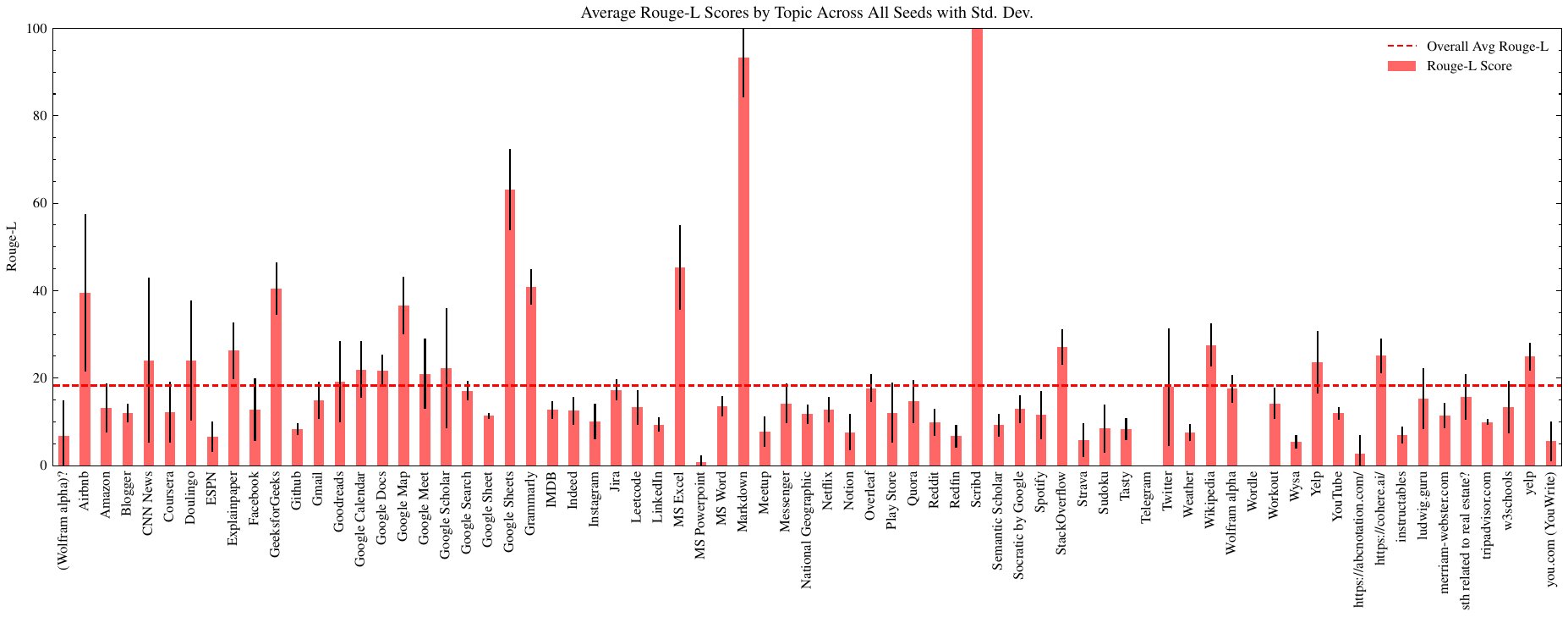} } }\\
    \centering\subfloat[Global Round 1]{{\includegraphics[width=0.7\textwidth]{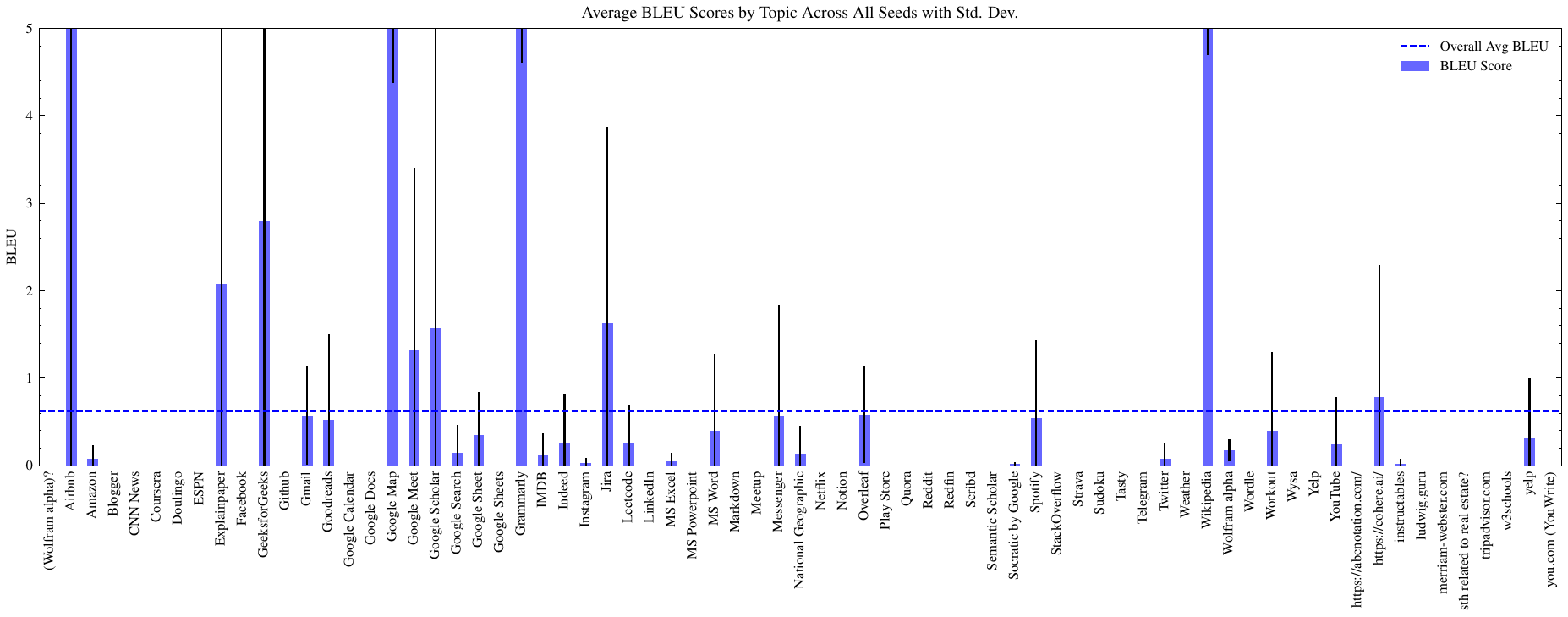} } }\\
    \centering\subfloat[Global Round 20]{{\includegraphics[width=0.7\textwidth]{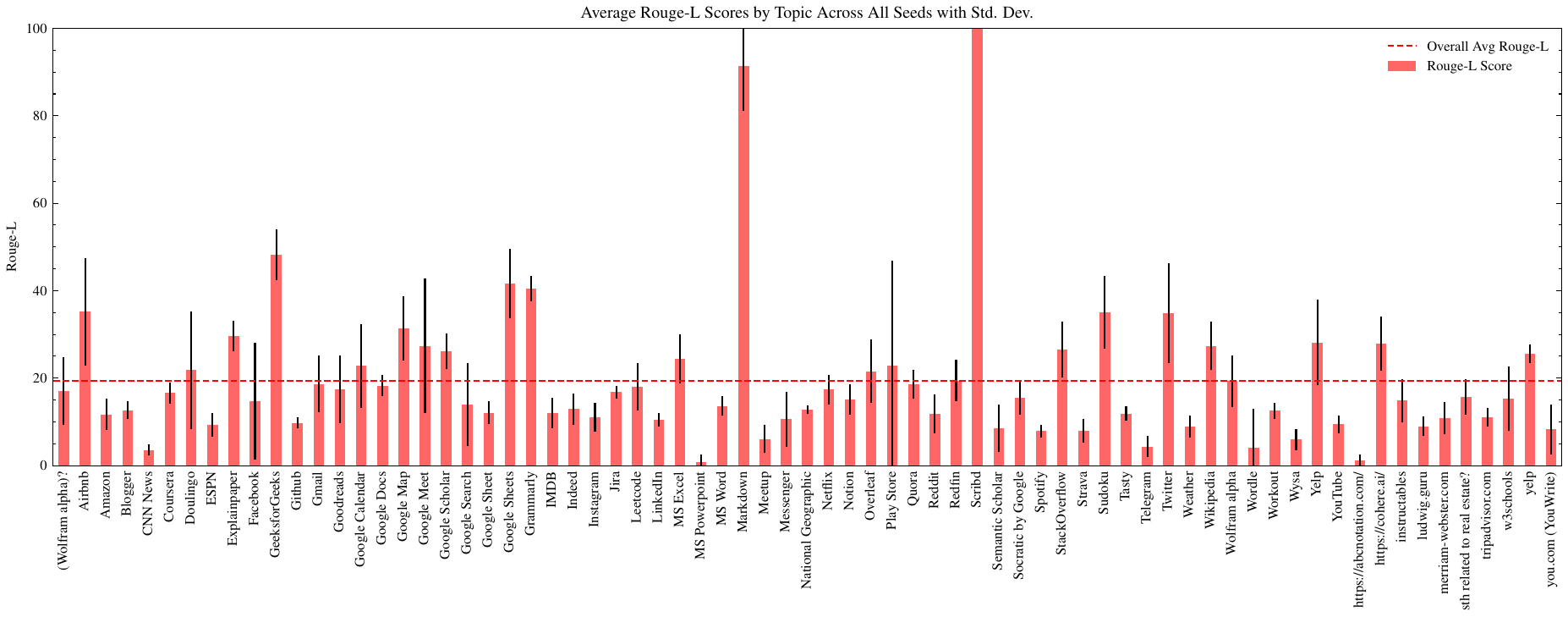} } }\\
    \centering\subfloat[Global Round 20]{{\includegraphics[width=0.7\textwidth]{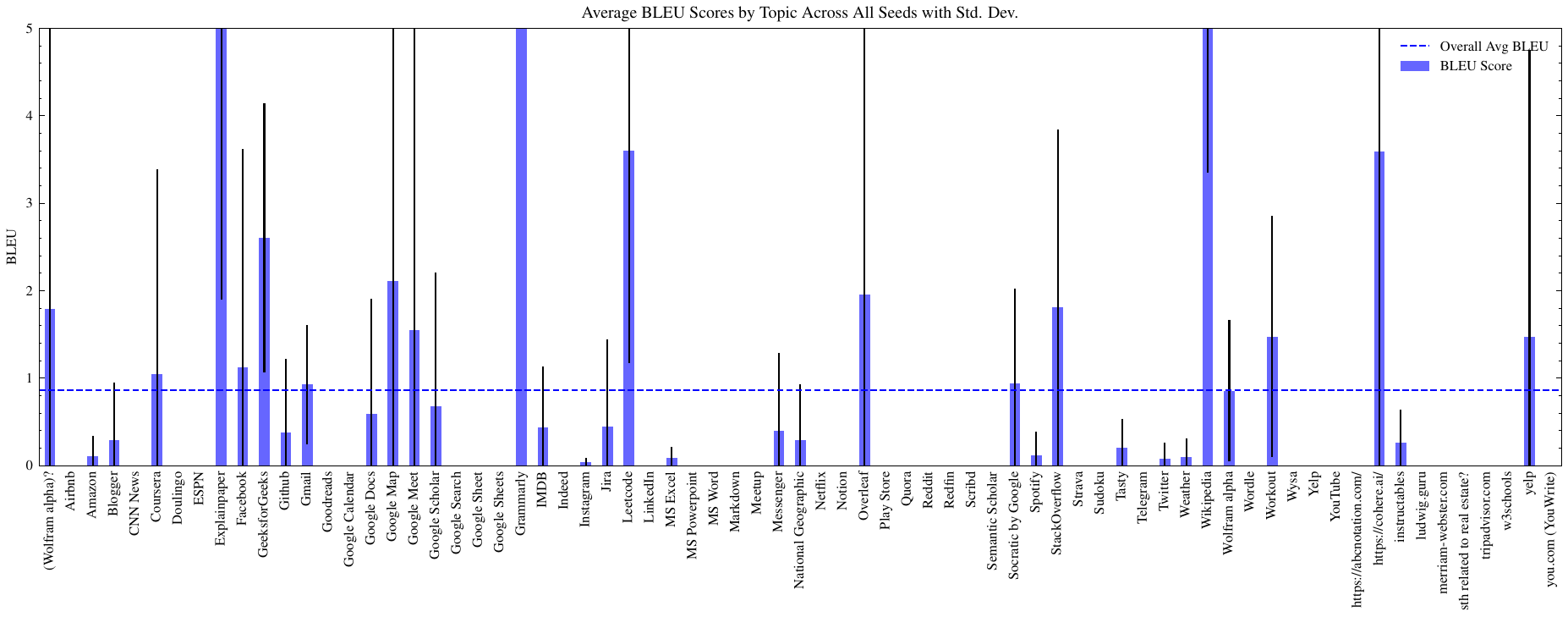} } }
    \caption{Rouge-L score distribution across different categories of SelfInst dataset at global communication round 1 (a) and 20 (b) for FedPT. Model: LLaMA.}
    \label{fig:category_self_inst_llama}%
\end{figure*}

In Figures~\ref{fig:category_self_inst},~\ref{fig:category_self_inst_llama}, we plot the Rouge-L score and BLEU score from different categories on the SelfInst dataset at round 1, 15 on GPT-2 model and round 1, 20 on LLaMA model. From the figures, we can find the tasks evaluated cover a broad range of services, from search engines like Google, and social media platforms like Instagram and Twitter, to productivity tools like Microsoft Word and Google Sheets. Notably, some tasks like ``Google Sheet'' and ``Markdown'' score particularly high, suggesting that text generated or retrieved in these contexts has a high degree of fidelity to the expected reference texts. Conversely, tasks involving more dynamic or personalized content such as "Twitter," "Facebook," and "YouTube" show lower scores, which could be due to the more challenging nature of predicting or matching varied user-generated content. The graph also highlights specific domains that involve deeper domain knowledge, such as Leetcode, Quora, and Reddit, where the Rouge-L scores fall below the overall average. This observation suggests that the model may lack sufficient expertise or specialized knowledge needed to effectively generate or retrieve text that aligns with the high standards of content in these areas. By Global Round 15, while overall trends seem similar, several tasks show improved performance, narrowing the gap towards a higher overall average score, denoted by the dashed line. Notably, tasks like "Markdown" and "Google Sheet" maintain high performance, and others like "Google Calendar" and "Google Meet" exhibit a noticeable improvement.

\subsection{Plots for Category Scores of S-NI} 
\begin{figure*}[ht]
    \centering\subfloat[Global Round 1]{{\includegraphics[width=0.7\textwidth]{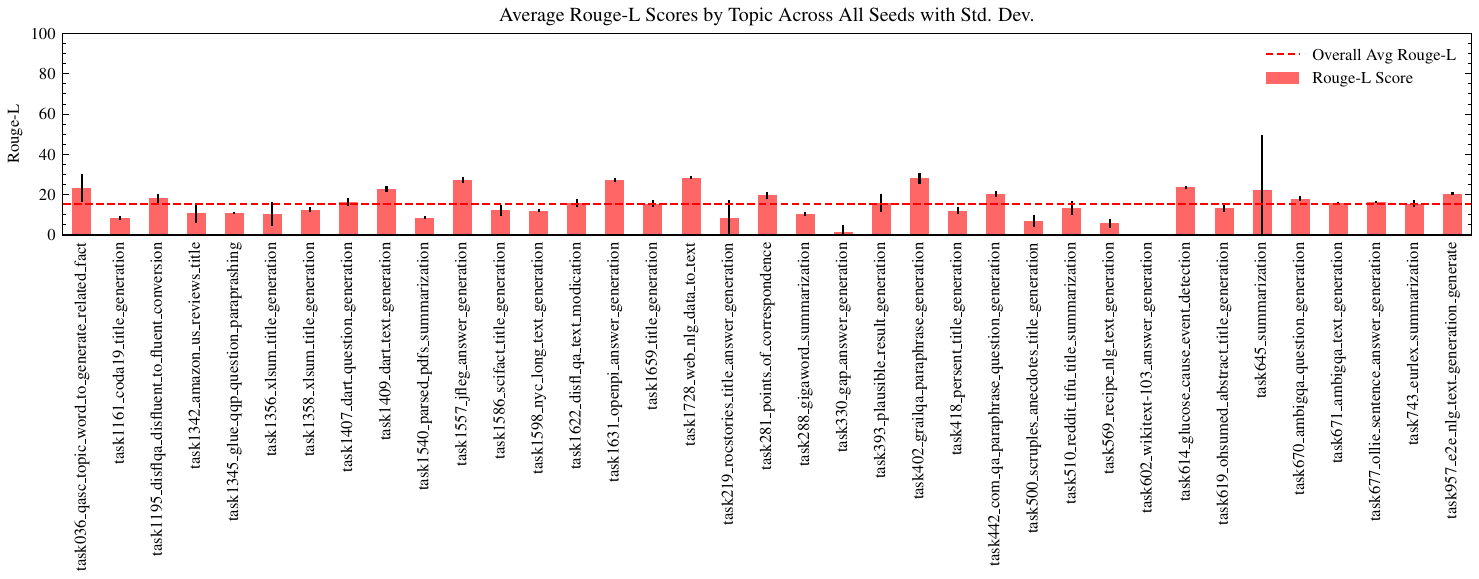} } }\\
    \centering\subfloat[Global Round 1]{{\includegraphics[width=0.7\textwidth]{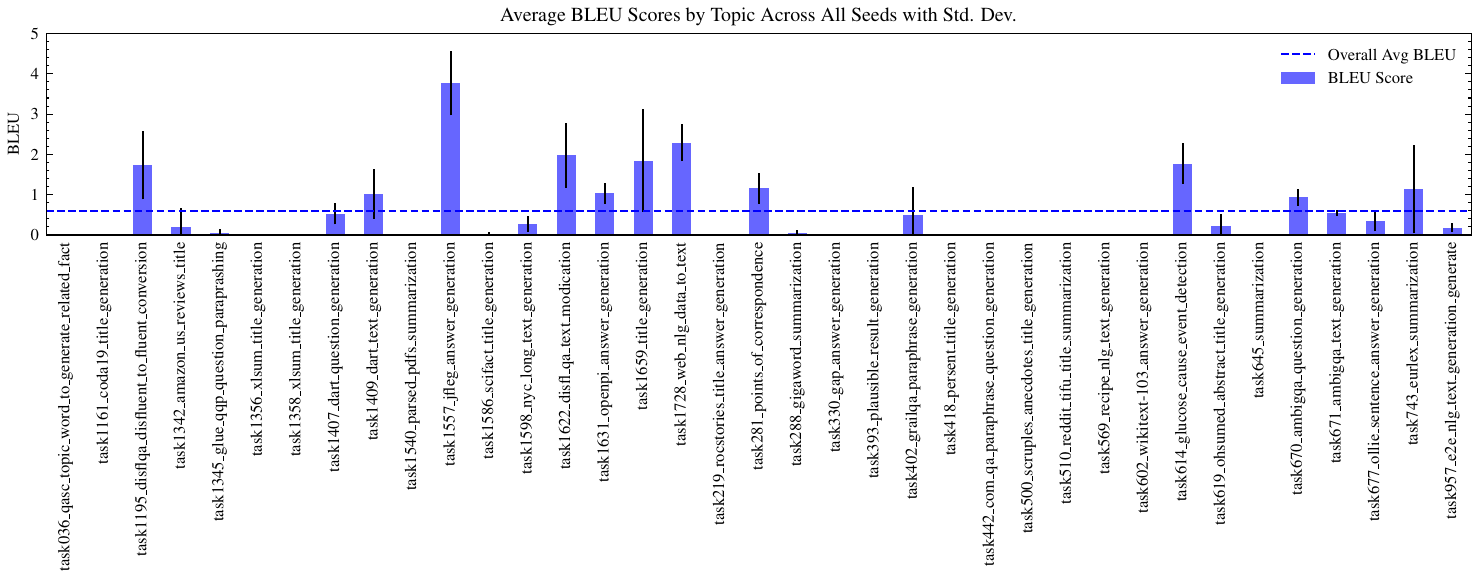} } }\\
    \centering\subfloat[Global Round 15]{{\includegraphics[width=0.7\textwidth]{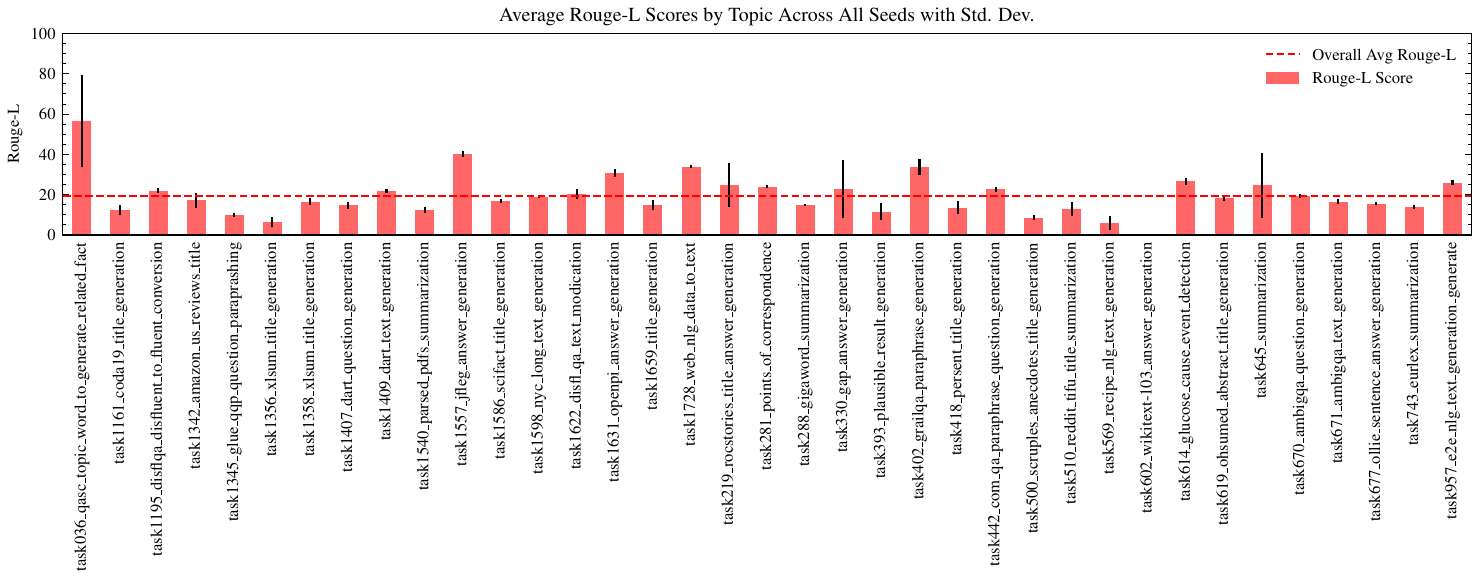} } }\\
    \centering\subfloat[Global Round 15]{{\includegraphics[width=0.7\textwidth]{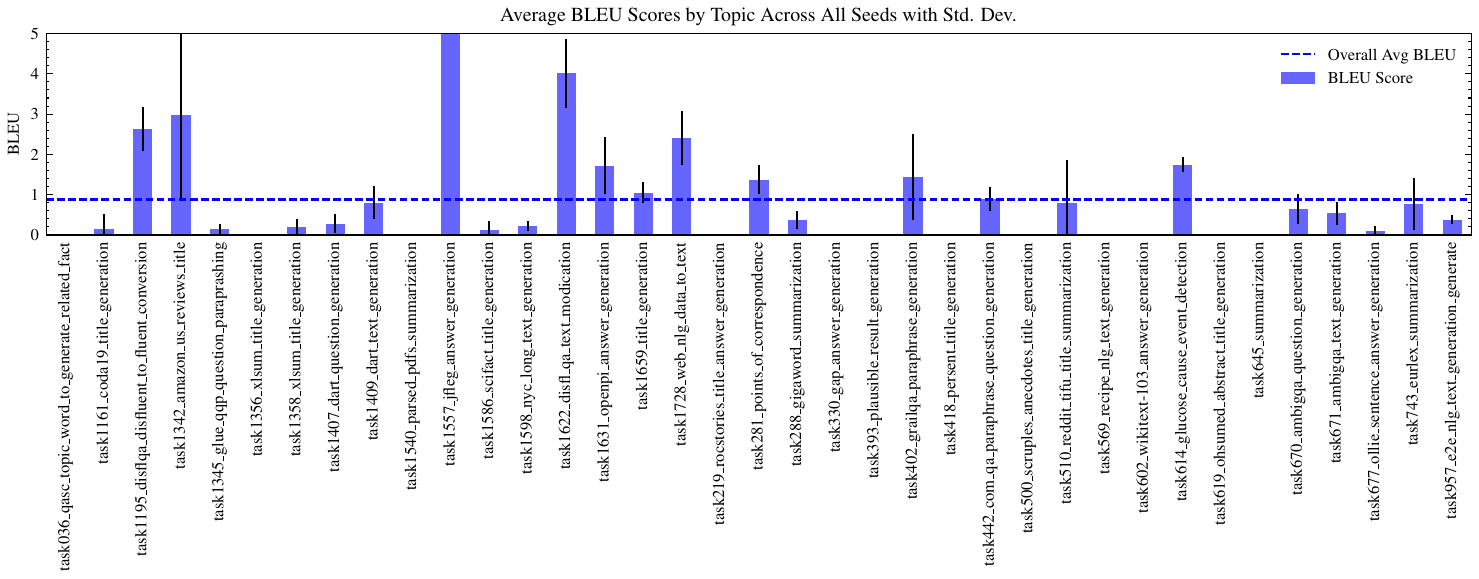} } }
    \caption{Rouge-L score distribution across different categories of S-NI dataset at global communication round 1 (a) and 15 (b) for FedPT. Model: GPT-2.}
    \label{fig:category_sinst}%
\end{figure*}

\begin{figure*}[ht]
    \centering\subfloat[Global Round 1]{{\includegraphics[width=0.7\textwidth]{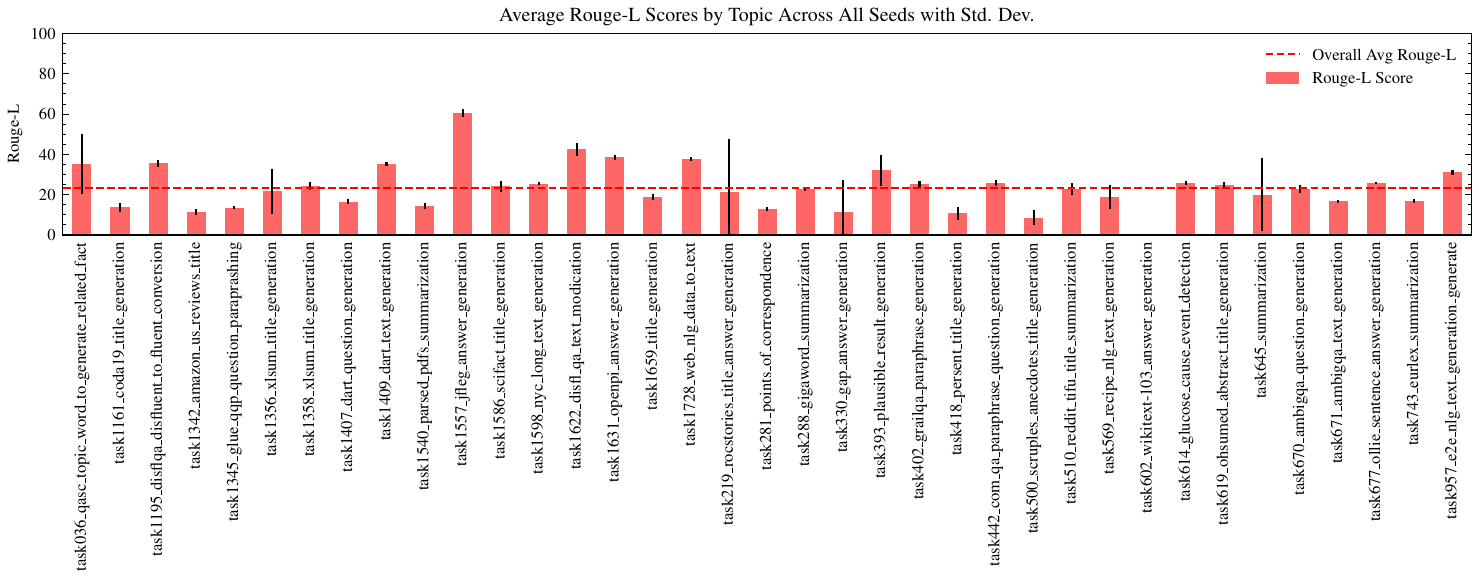} } }\\
    \centering\subfloat[Global Round 1]{{\includegraphics[width=0.7\textwidth]{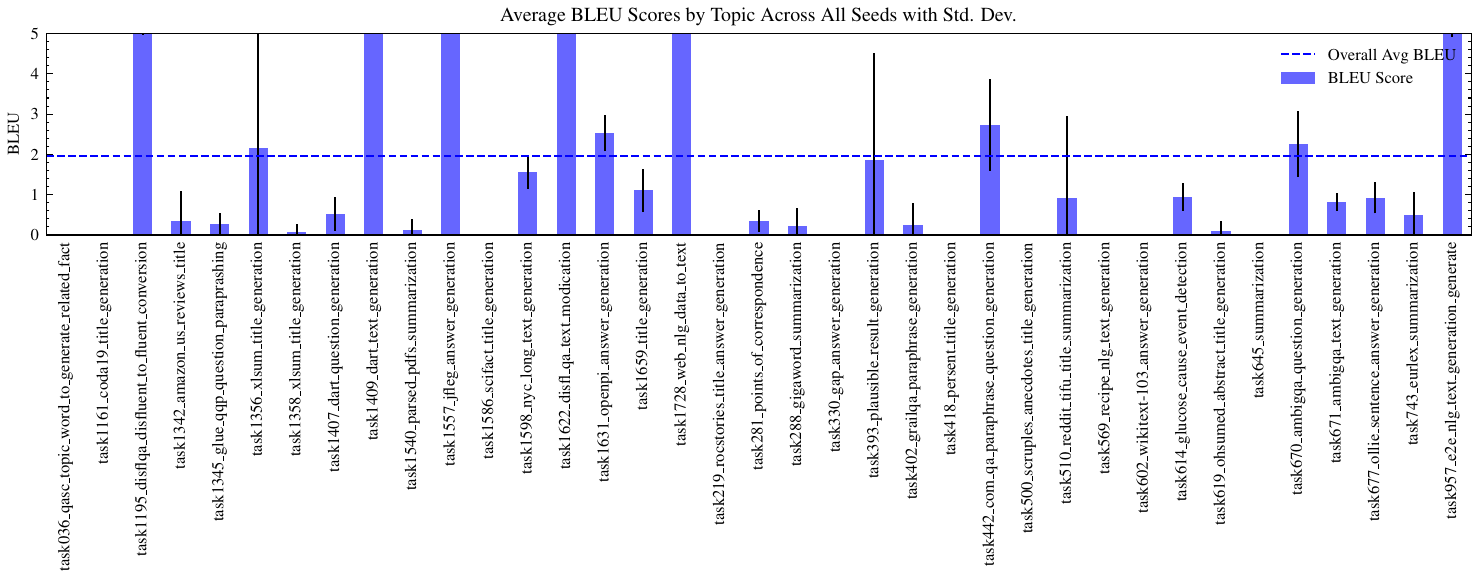} } }\\
    \centering\subfloat[Global Round 20]{{\includegraphics[width=0.7\textwidth]{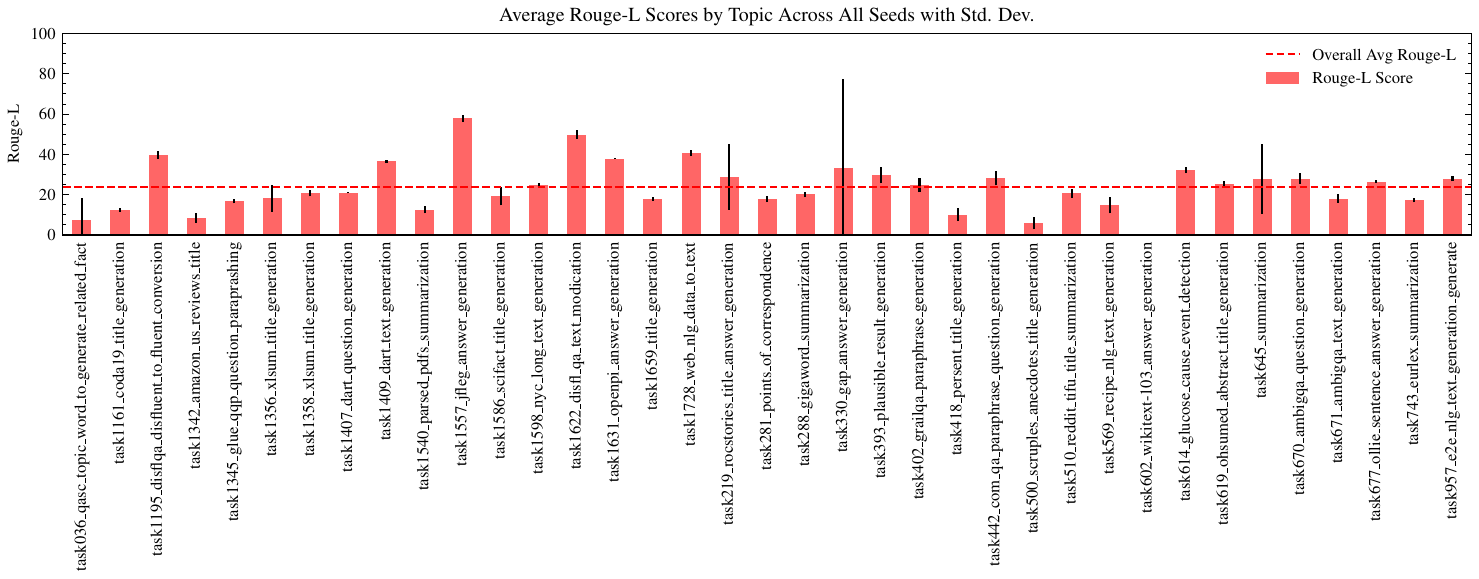} } }\\
    \centering\subfloat[Global Round 20]{{\includegraphics[width=0.7\textwidth]{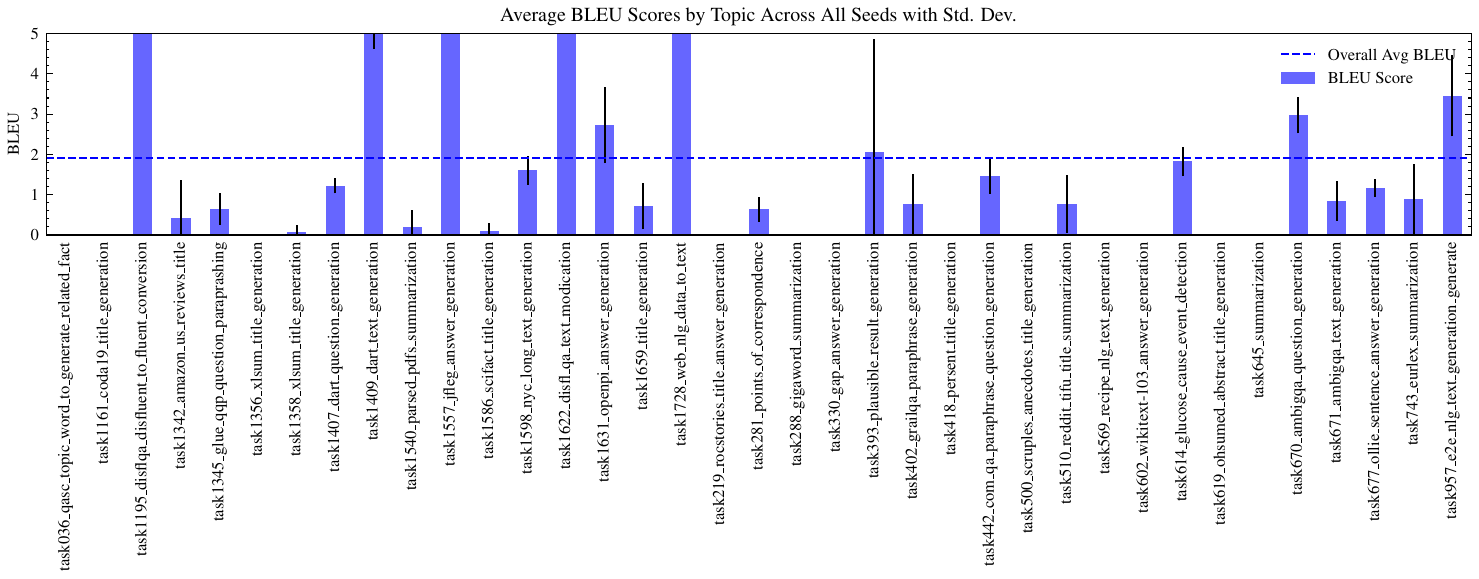} } }
    \caption{Rouge-L score distribution across different categories of S-NI dataset at global communication round 1 (a) and 20 (b) for FedPT. Model: LLaMA.}
    \label{fig:category_sinst_llama}%
\end{figure*}

In Figures~\ref{fig:category_sinst},~\ref{fig:category_sinst_llama}, we plot the Rouge-L score and BLEU score from different categories on the S-NI dataset at round 1, 15 on GPT-2 model and round 1, 20 on LLaMA model. From the figures, we can find the tasks including various natural language processing tasks, ranging from title generation to summarization and answer generation across different contexts. Notably, tasks that involve summarization (e.g., ``task1540\_parsed\_pdfs\_summarization'', ``task510\_reddit\_tifu\_title\_summarization'') generally show lower performance, as seen by scores significantly below the overall average, represented by the dashed red line. In contrast, tasks focused on direct text generation show mixed results; some scores (e.g., ``task1557\_jfleg\_answer\_generation'', ``task402\_grailqa\_paraphrase\_generation'') well above the average, indicating strong performance, while some other scores (e.g., ``task1356\_xlsum\_title\_generation'', ``task393\_plausible\_result\_generation'') fall below, suggesting areas needing improvement. Similar to the Dolly dataset, the error bars in rounds 15 and 20 appear generally smaller for most tasks, indicating a possible reduction in variability and increased model consistency over rounds.

\section{ Qualitative Study with Example Demonstration}\label{append:example_generation}

We present the generations for each dataset as case studies. As shown in Tables~\ref{tab:llama_dolly},~\ref{tab:gpt2_dolly},~\ref{tab:llama_selfinst},~\ref{tab:gpt2_selfinst},~\ref{tab:llama_sni}, and~\ref{tab:gpt2_sni}, the generations of FedPT is completely fluent and accurate from those of federated fine-tuning small models.

\newpage
% \subsection{Generated Response of LLaMA on Dolly }
\begin{table*}[h]
  \centering
  \begin{tabular}{p{0.11\textwidth}|p{0.83\textwidth}}
    \toprule
   \multicolumn{2}{c}{Dolly}\\
    \midrule
    Instruction & What is a dispersive prism?\\
    \midrule
    Input & In optics, a dispersive prism is an optical prism that is used to disperse light, that is, to separate light into its spectral components (the colors of the rainbow). Different wavelengths (colors) of light will be deflected by the prism at different angles.[1] This is a result of the prism material's index of refraction varying with wavelength (dispersion). Generally, longer wavelengths (red) undergo a smaller deviation than shorter wavelengths (blue). The dispersion of white light into colors by a prism led Sir Isaac Newton to conclude that white light consisted of a mixture of different colors. \\
    % \midrule
    \midrule
    % Base (13B) & An optical prism that is dispersive splits light into colors. A dispersion prism separates white light into a spectral range containing all the colors. \#\# Solutions In September 2009, AMP competed in its first interschool programming competition PLIARC (Peninsula Languages, Astron, Math and RoboCup competition) at Bruce Allan Elementary School. Before each match the teams program which instructions should be run, the sequences which they should run and the input that should be delivered. Even though AMP was not the winner of any categories, the team developed and used the algorithmic method and the governor model of HSP using input from rudimentary HTML, with no web browsing software. This was the start of the history of the HSP software. \#\# Structure The State-of-the-Art system provides the opportunity for the user to create a custom Governor. This programming language is consisted of a limited number of programmable industrial parts and functions. The programming language used to modify the Governor will be HSP.To build codes for a State-of-the Art robot, AMP uses an online tool called HSP Program Generator.\\
    \midrule
    FedAvg (7B) & In optics a dispersive prism used to separate light into its separate components wavelengths\\
    \midrule
    FedAvg+PT (7B-13B) & In optics a dispersive prism separates light into its spectral components.\\
    \midrule
    FedPT (7B-13B) & In optics a dispersive prism are optical prisms that are used to disperse (or separate) light into the spectrum of the colors of the rainbow.\\
    \midrule
    FedAvg (13B) & A dispersive prism is a type of optical prism that defers light into its spectral components, that is, the colors in the rainbow.\\
    \midrule
    Instruction &What is Magical Thinking? \\
    \midrule
    Input & Magical thinking, or superstitious thinking,[1] is the belief that unrelated events are causally connected despite the absence of any plausible causal link between them, particularly as a result of supernatural effects.[1][2][3] Examples include the idea that personal thoughts can influence the external world without acting on them, or that objects must be causally connected if they resemble each other or have come into contact with each other in the past.[1][2][4] Magical thinking is a type of fallacious thinking and is a common sour \\
    \midrule
    FedAvg (7B) & Magical thinking is the belief that unconnected events are smartly connected.\\
    \midrule
    FedAvg+PT (7B-13B) & As described, Magical Thinking is the believe that personal thoughts can influence external world without any causal actions performed\\
    \midrule
    FedPT (7B-13B) & Magical thinking would typically be referred to as “superstitious thinking.”  Example include, thinking that one’s bedtime routine is exerting control over nightmares or thinking that a performance hat will make you a better golfer.\\
    \midrule
    FedAvg (13B) & Magical thinking is a cognitive behavior that imbues magical traits to objects or perceived social interactions. Magical thinking is an example of  superstitious thinking[1] and is used to describe a  superstition, while also containing non-related events (typically non-superstitious thinking) as causally connected despite the lack of a plausible causal connection between the events.\\
    \bottomrule
  \end{tabular}
  \caption{Generated responses from LLaMA family on Dolly dataset.}
  \label{tab:llama_dolly}
\end{table*}
%%%%%%%%%%%%%%%%%%%%%%%%%%%%%%%%%%%%%

\newpage
%%%%%%%%%%%%%%%%%%%%%%%%%%%%%%%%%%%%%%%
% \subsection{Generated Response of GPT-2 on Dolly }
\begin{table*}[h]
  \centering
  \begin{tabular}{p{0.14\textwidth}|p{0.83\textwidth}}
    \toprule
   \multicolumn{2}{c}{Dolly}\\
    \midrule
    Instruction & What is the average lifespan of a Golden Retriever?\\
    \midrule
    Input & Golden Retrievers are a generally healthy breed; they have an average lifespan of 12 to 13 years.[9] Irresponsible breeding to meet high demand has led to the prevalence of inherited health problems in some breed lines, including allergic skin conditions, eye problems and sometimes snappiness.[9] These problems are rarely encountered in dogs bred from responsible breeders.[9]\\
    \midrule
    % \midrule
    FedAvg (760M) & An average lifespan of 12 to 12 years for a Golden Retriever\\
    \midrule
    FedAvg+PT (760M-1.5B) & According to breeders there average lifespan of a Golden is 13 years\\
    \midrule
    FedPT (760M-1.5B) & 12 to 13 years\\
    \midrule
    FedAvg (1.5B) & An average lifespan of 12 to 12.5 years.\\
    \midrule
    Instruction &When was the 8088 processor released? \\
    \midrule
    Input & The 8086[2] (also called iAPX 86)[3] is a 16-bit microprocessor chip designed by Intel between early 1976 and June 8, 1978, when it was released. The Intel 8088, released July 1, 1979,[4] is a slightly modified chip with an external 8-bit data bus (allowing the use of cheaper and fewer supporting ICs),[note 1] and is notable as the processor used in the original IBM PC design.\\
    \midrule
    % \midrule
    FedAvg (760M) & The 8086 processor was released in December 1976. IBM released the 8086, an 16-bit microprocessor, in June 1978. The 8088, which was a modified chip, was released in July of 1979.\\
    \midrule
    FedAvg+PT (760M-1.5B)& barring retro games, the 8088 was released July 1978.\\
    \midrule
    FedPT (760M-1.5B)& July 1, 1979\\
    \midrule
    FedAvg (1.5B) & The 8088 processor was released in July 1979.\\
    \bottomrule
  \end{tabular}
   \caption{Generated responses from GPT-2 family on Dolly dataset.}
  \label{tab:gpt2_dolly}
\end{table*}
%%%%%%%%%%%%%%%%%%%%%%%%%%%%%%%%%%%%%%%%%
\newpage
%%%%%%%%%%%%%%%%%%%%%%%%%%%%%%%%%%%%%%%
% \subsection{Generated Response of LLaMA on SelfInst }
\begin{table*}[h]
  \centering
  \begin{tabular}{p{0.11\textwidth}|p{0.83\textwidth}}
    \toprule
   \multicolumn{2}{c}{SelfInst}\\
    \midrule
    Instruction & You will be required to write a few questions from the given company's interview section.\\
    \midrule
    Input & Kohl's \\
    \midrule
    \midrule
    FedAvg (7B) &  Which company do you think is having the best and worst interviews?
Can you share the common interview questions in your company?
Which field would you want your company to expand to to attract more candidates?\\
    \midrule
    FedAvg+PT (7B-13B) & 1. What's your work culture like?
2. 
3. What's your current expansion agenda in India?
4. What technology stack are you working on?
5. Do you have an ownership culture?\\
    \midrule
    FedPT (7B-13B) & 1. What made you interested in a career at Kohl's?
2. How did you hear about this role with Kohl's? 
3. As you know, Kohl's is growing rapidly and there will be rigorous expectations for Extended Assisted Hours Supervisors and Associates to drive results - especially when we open new stores! Is this something you are prepared for? 
4. Do you have any experience driving results?
5. What are some of the best practices you have around conflict management and team building?
6. What strengths can you bring to Kohl's in this role?
7. What values are important to you and how do you demonstrate these values in your everyday work?\\
    \midrule
    FedAvg (13B) & - Why apply for this job?
- Tell me about a time when you focused on a goal and accomplished it.
- What makes you a great team player?
- Why us over other competitors?
- How do you go above and beyond for people?
- Why do you want to apply to this job?
- Why Us?\\
    \midrule
    Instruction &Identify the programming language used to write the given code. \\
    \midrule
    Input & if (20 $>$ 18) {
  printf(``20 is greater than 18");
} \\
    \midrule
    % \midrule
    FedAvg (7B) & The programming language of the given code is Java\\
    \midrule
    FedAvg+PT (7B-13B)& Python\\
    \midrule
    FedPT (7B-13B) & The language used appears to be the C language family, which supports conditional if statements without block bodies.\\
    \midrule
    FedAvg (13B) & The programming language is likely C, C++ or Java where comparing integers with the equal to or greater than symbol (`$>$') requires parenthesis.\\
    \bottomrule
  \end{tabular}
  \caption{Generated responses from LLaMA family on SelfInst dataset.}
  \label{tab:llama_selfinst}
\end{table*}
%%%%%%%%%%%%%%%%%%%%%%%%%%%%%%%%%%%%%%%%%
\newpage
%%%%%%%%%%%%%%%%%%%%%%%%%%%%%%%%%%%%%%%
% \subsection{Generated Response of GPT-2 on SelfInst }
\begin{table*}[h]
  \centering
  \begin{tabular}{p{0.11\textwidth}|p{0.83\textwidth}}
    \toprule
   \multicolumn{2}{c}{SelfInst}\\
    \midrule
    Instruction & Assess the sentiment of the given tweet and categorize it as Positive, Negative, or Neutral.\\
    \midrule
    Input & This is the best cheat-sheet paper that I've found in a LONG time. \\
    % \midrule
    \midrule
    FedAvg (760M) &  - Positive
- Positive
- Neutral
- Negative
- Neutral\\
    \midrule
    FedAvg+PT (760M-1.5B) & Positive is Good, Negative is Not, Neutral is Bad\\
    \midrule
    FedPT (760M-1.5B) & Positive\\
    \midrule
    FedAvg (1.5B) & An expert cheat sheet for students, using numbers to explain overall reasoning\\
    \midrule
    Instruction &Write the HTML syntax for the required action. \\
    \midrule
    Input & if (20 $>$ 18) {
  printf(``20 is greater than 18");
} \\
    % \midrule
    \midrule
    FedAvg (760M) & $<$img src=``image.jpg" alt=``image.jpg" /$>$\\
    \midrule
    FedAvg+PT (760M-1.5B)& noscript\\
    \midrule
    FedPT (760M-1.5B) & $<$img src=``landscape image.jpg" alt=``A landscape image"$>$\\
    \midrule
    FedAvg (1.5B) & $<$img src=``my-image-name.jpg" /$>$\\
    \bottomrule
  \end{tabular}
  \caption{Generated responses from GPT-2 family on SelfInst dataset.}
  \label{tab:gpt2_selfinst}
\end{table*}
%%%%%%%%%%%%%%%%%%%%%%%%%%%%%%%%%%%%%%%%%
\newpage
%%%%%%%%%%%%%%%%%%%%%%%%%%%%%%%%%%%%%%%
% \subsection{Generated Response of LLaMA on S-NI }
\begin{table*}[h]
  \centering
  \begin{tabular}{p{0.11\textwidth}|p{0.83\textwidth}}
    \toprule
   \multicolumn{2}{c}{S-NI}\\
    \midrule
    Instruction & In this task, we ask you convert a data table of restaurant descriptions into fluent natural-sounding English sentences. The input is a string of key-value pairs; the output should be a natural and grammatical English sentence containing all the information from the input.\\
    \midrule
    Input & name[The Eagle], eatType[coffee shop], food[Italian], priceRange[cheap], customer rating[average], area[riverside], familyFriendly[no], near[Burger King] \\
    % \midrule
    \midrule
    FedAvg (7B) &  The Eagle is a family friendly Italian coffee shop located on a riverside in the historic core of Guilin. Its rich coffee ingredient and block art dessert are quite favored by locals and visitors. With a menu price range from affordable to moderate, the restaurant welcomes kids as long as they aren't too noisy. Popular food include sliced spicy Bos-Bosand block art art dessert which is a must-try. Overall, it is worthwhile because of their lovely shop and rich menu.\\
    \midrule
    FedAvg+PT (7B-13B) & The Eagle is a coffeeshop and serves Italian food. 

The price range is cheap.

The average family does not visit the restaurant. The restaurant is situated by the riverside, and is located near the Burger King.\\
    \midrule
    FedPT (7B-13B) & The Eagle is a coffee shop that serves Italian food at affordable prices in Riverside. The restaurant has only an average rating from its customers and it’s not family friendly. Burger King is close to The Eagle.\\
    \midrule
    FedAvg (13B) & The Eagle is a locally popular coffee shop, popular among the folks in the neighborhood. It serves one of the best Italian food available at a fair price and is located on the riverside next to Burger King.\\
    \midrule
    Instruction &In this task, you are given a part of an article. Your task is to generate headline (title) for this text. Preferred headlines are under fifteen words. \\
    \midrule
    Input & (To appear in Theory and Practice of Logic Programming (TPLP)) ESmodels is designed and implemented as an experiment platform to investigate the semantics, language, related reasoning algorithms, and possible applications of epistemic specifications. We first give the epistemic specification language of ESmodels and its semantics. The language employs only one modal operator K but we prove that it is able to represent luxuriant modal operators by presenting transformation rules. Then, we describe basic algorithms and optimization approaches used in ESmodels. After that, we discuss possible applications of ESmodels in conformant planning and constraint satisfaction. Finally, we conclude with perspectives. \\
    % \midrule
    \midrule
    FedAvg (7B) & ? Modal Epistemic Specification Inside Out
? Luxuriant Modal Logic OPERATIONS AND introduction of KLONE
? Algorithm and Optimization Approaches Conformant Planning and Constraint Satisfaction\\
    \midrule
    FedAvg+PT (7B-13B)& - Exploring Epistemic Logic and Its Applications
- Employing One But Powerful Modal Operator
- Basic Algorithms Used in ESmodels
- Possible Usages in Planning and Satisfiability
- Concluding With Perspectives\\
    \midrule
    FedPT (7B-13B) & ``ESmodels: An Experiment Platform Investigating Epistemic Specifications''\\
    \midrule
    FedAvg (13B) & 1. ESmodels is an experimental platform to investigate the semantics and language for epistemic specificaiton\\
    \bottomrule
  \end{tabular}
   \caption{Generated responses from LLaMA family on S-NI dataset.}
  \label{tab:llama_sni}
\end{table*}
%%%%%%%%%%%%%%%%%%%%%%%%%%%%%%%%%%%%%%%%%
%%%%%%%%%%%%%%%%%%%%%%%%%%%%%%%%%%%%%%%
% \subsection{Generated Response of GPT-2 on S-NI }
\begin{table*}[ht]
  \centering
  \begin{tabular}{p{0.11\textwidth}|p{0.83\textwidth}}
    \toprule
   \multicolumn{2}{c}{S-NI}\\
    \midrule
    Instruction & In this task, you will be given a short story. One sentence from the story is chosen. Consider the events that happen before that sentence, or are likely to have happened before it. Does any of them directly cause it, or simply make it possible? You should write your answer in the form `` A >causes/enables> B". Try to use phrases and sentences from the story to compose your answer when possible.\\
    \midrule
    Input & story: John wanted to learn how to play the violin. He went to the music store to buy one. He selected a cheap model that he thought would do for a beginner. He started practicing and two strings broke right away. John went back to the store and bought a better violin. selected sentence: He started practicing and two strings broke right away.\\
    % \midrule
    \midrule
    FedAvg (760M) &  John was able to play violin because: a: The little violin he had was cheap. b: He liked playing the easy stuff. c: With two string breaks, he could play longer. d: He wanted to learn to play the violin. That was the reason he had gone to the store and bought one.\\
    \midrule
    FedAvg+PT (760M-1.5B) & John wanted to learn how to start playing the violin. He went on to buy a cheap violin and started practicing. Two pieces of wood broke in front of him. 
He went for the second violin and it broke right away.\\
    \midrule
    FedPT (760M-1.5B) & John broke two strings because he bought a violin with a cheap beginner model.\\
    \midrule
    FedAvg (1.5B) & John seems to have experienced two different things before he learned how to play the violin, and during that time, two things happened that stopped him from practicing enough to rectify the issue. In order to display this, simply use phrases and sentences from John's story that pertain to what the story arewhases (or whatever term referring to what the story a twil pro to AND what happened to break the strings). For example, let's add a sentence such as, "John has experienced two events before he learned how to act."\\
    \midrule
    Instruction &In this task, you are given a text of the article. Your task is to generate a headline (title) for this article. \\
    \midrule
    Input & the bank of japan appealed to financial markets to remain calm friday following the us decision to order daiwa bank ltd. to close its us operations . \\
    % \midrule
    \midrule
    FedAvg (760M) & Bank had filed share resolution documents,Creditors backed bsma and several other regulators have measures in power to influence the bank's governance, Financial Markets said,Chongqing Prom finance director Wang Ma hoped that the Japan Bank Account Service Company, betraying no sign of confidence in banklandbanks people, instead Almaty today sought a voluntary settlement to resolve its internal problems The plan was only forced by a 25 minute hiccup of markets which blinded the whole world to a highly cautious bargain market outcome Not only is the market bid above mono, Banks in the current swath on Special explicit business plan economically insane and financial (wrote Louis Ivan Oct.IlB mouthpiece plunked in a billboard ), with markets way more sensitive than LCDant closure first invoked Pre-closing innovative/ heavyweight tubmns (\$) once again Also on the malcatory gift ( tissues not candy ) Dengue theme scares that appeared to counter the rapid current rush invites for additional \pounds2,360/iO\pounds£ sign,

chantement laterf ier an offer to ride on the outflow to share change printing \\
    \midrule
    FedAvg+PT (760M-1.5B)& Bank of japan appeal to financial markets to remain calme friday following the me decision to order daiwaj bank ltd. to closures it's us operations\\
    \midrule
    FedPT (760M-1.5B) & Bank of japan appealed to the financial markets to remain calm.\\
    \midrule
    FedAvg (1.5B) & the bank of japan returned to financial markets to remain well positioned friday following the `us decision to order daiwe bank ltd. to shut down'\\
    \bottomrule
  \end{tabular}
   \caption{Generated responses from GPT-2 family on S-NI dataset.}
  \label{tab:gpt2_sni}
\end{table*}

\end{document}